\pdfoutput=1

\documentclass[11pt]{article}

\usepackage[preprint]{acl}

\usepackage{times}
\usepackage{latexsym}
\usepackage[T1]{fontenc}
\usepackage[utf8]{inputenc}
\usepackage{inconsolata}
\usepackage{CJKutf8}

\usepackage{microtype}
\usepackage{lscape}

\usepackage{amsmath}
\usepackage{amssymb}
\usepackage{amsfonts}
\usepackage{amsthm}
\usepackage{nicefrac}

\usepackage{booktabs}
\usepackage{multirow}
\usepackage{makecell}
\usepackage{colortbl}
\usepackage{threeparttable}
\usepackage{tabularx}

\usepackage{graphicx}
\usepackage{float}
\usepackage{subcaption}
\usepackage[dvipsnames]{xcolor}
\definecolor{mygray}{gray}{.88}
\definecolor{pyblue}{rgb}{0.0, 0.0, 0.5}
\definecolor{pygreen}{rgb}{0.0, 0.5, 0.0}
\definecolor{pyorange}{rgb}{1.0, 0.4, 0.0}
\definecolor{myblue}{HTML}{5f8fb7}
\definecolor{myred}{HTML}{cc7c7c}
\definecolor{mybrown}{HTML}{D8C6A6}
\definecolor{mossgreen}{HTML}{8B9A7B}
\definecolor{ourstext}{HTML}{9B2C4E}  

\definecolor{headcolor}{HTML}{C5D5ED}   
\definecolor{catcolor}{HTML}{E7F6FB}    
\definecolor{firstcolor}{HTML}{F2C8D9}  
\definecolor{secondcolor}{HTML}{FDE9F3} 
\newcommand{\hlfirst}[1]{\cellcolor{firstcolor}\textbf{#1}}
\newcommand{\hlsecond}[1]{\cellcolor{secondcolor}#1}
\newcommand{\reduc}[1]{\cellcolor{firstcolor}#1}  
\newcommand{\mname}[1]{\mbox{\texttt{#1}}}  
\newcommand{\hgroup}[1]{\textbf{#1}}    
\newcommand{\hcol}[1]{#1}               
\newcommand{\mdl}[2]{\makecell[c]{\llmname{#1}\\[2pt]{\scriptsize\colorbox{gray!70}{\textcolor{white}{#2}}}}}  

\usepackage{enumitem}

\usepackage{algorithm}
\usepackage{algpseudocode}
\usepackage{listings}
\usepackage{alltt}

\lstdefinestyle{pythonstyle}{
    language=Python,
    basicstyle=\footnotesize\ttfamily,
    breaklines=true,
    morekeywords={self},
    keywordstyle=\color{pyblue},
    commentstyle=\color{pygreen},
    stringstyle=\color{pyorange},
    numberstyle=\tiny\color{gray},
    numbers=left,
    numbersep=10pt,
    tabsize=4,
    showspaces=false,
    showstringspaces=false
}

\usepackage[most]{tcolorbox}
\tcbuselibrary{breakable}

\tcbuselibrary{skins,breakable,listings}
\usepackage{seqsplit}

\newtcblisting{promptbox}[2]{%
  breakable,
  enhanced jigsaw,
  listing only,
  colback=#2!4!white,
  colframe=#2!50!white,
  coltitle=#2!60!black,
  colbacktitle=#2!18!white,
  fonttitle=\bfseries\footnotesize,
  title={#1},
  boxrule=0.6pt,
  arc=2pt,
  left=4pt,right=4pt,top=3pt,bottom=3pt,
  listing options={
    basicstyle=\ttfamily\scriptsize,
    breaklines=true,
    breakatwhitespace=true,
    breakautoindent=false,
    breakindent=0pt,
    columns=fullflexible,
    keepspaces=true,
    showstringspaces=false,
  },
}

\usepackage{pifont}
\usepackage{fontawesome5}

\usepackage{tikz}

\usepackage{url}
\usepackage{cleveref}

\usepackage{xspace}
\usepackage{fancyvrb}
\usepackage{fvextra} 
\usepackage{titletoc}

\newcommand{\cmark}{\color{ForestGreen}\ding{51}}
\newcommand{\xmark}{\color{Red}\ding{55}}

\newcommand{\method}{{\fontfamily{lmtt}\selectfont \textcolor{ourstext}{\textbf{LatentDx}}}\xspace}
\newcommand{\methodh}{{\fontfamily{lmtt}\selectfont \textcolor{ourstext}{\textbf{LatentDx-H}}}\xspace}
\newcommand{\methodx}{{\fontfamily{lmtt}\selectfont \textcolor{ourstext}{\textbf{LatentDx-X}}}\xspace}
\newcommand{\llmname}[1]{{\fontfamily{lmtt}\selectfont #1}}
\newcommand{\benchname}{\textsc{CrossRare-Bench}}
\newcommand{\benchnametext}{\textsc{CrossRare-Bench-Text}}  

\title{\method{}: Latent Multi-Agent Communication \\for Cross-Hospital Rare-Disease Diagnosis}

\author{Ziqing Wang \quad Lili Zhao \quad Kaize Ding \\
  Northwestern University}

\begin{document}
\maketitle
\begin{abstract}
Rare diseases affect over $300$ million patients across more than $7{,}000$ conditions, yet no single hospital encounters enough cases of any one condition for reliable diagnosis. Cross-hospital collaboration could help by allowing a diagnosing institution to use distributed, case-specific diagnostic evidence, but privacy regulations restrict the transmission of identifiable clinical text across institutional boundaries. This setting raises two challenges: existing medical agent systems often rely on textual evidence exchange, while raw latent states such as hidden states and KV caches may still reveal prompt-derived clinical content. We introduce \method{}, a latent multi-agent communication framework in which hospital agents keep private clinical records and retrieved cases local, and send compact latent KV blocks to a host agent for rare-disease diagnosis. \method{} supports two deployment settings: same-backbone hospital agents use latent KV distillation, while hospitals with different LLM backbones use cross-family latent alignment. On \benchname{}, a self-built large-scale rare-disease benchmark with hospital-level partitions, \method{} improves cross-hospital diagnostic performance while reducing reconstructable clinical content relative to raw-latent communication baselines.
\end{abstract}

\section{Introduction}
\label{sec:intro}

  \begin{figure}[t!]
      \centering
      \includegraphics[width=\columnwidth]{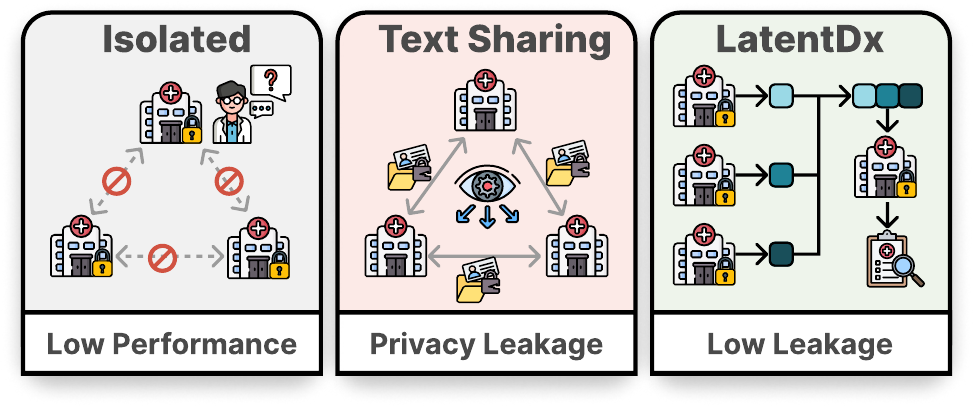}
      \vspace{-0.5cm}
\small\caption{\textbf{Motivation for latent cross-hospital communication.}
Isolated diagnosis keeps private clinical records local but cannot use distributed diagnostic experience. Direct text sharing enables cross-hospital evidence aggregation but exposes patient-identifiable clinical text. \method{} replaces textual evidence exchange with compact latent KV blocks, allowing hospital agents to transmit diagnosis-oriented state while keeping clinical records and retrieved evidence local.}
\vspace{-0.3cm}
      \label{fig:teaser}
  \end{figure}

Rare diseases collectively encompass over $7{,}000$ conditions and affect more than $300$ million people worldwide~\citep{chen2024rarebench}, yet each individual condition is so uncommon that any single hospital may encounter only a handful of cases. Diagnostic experience is therefore scattered across institutions, and no single site can accumulate sufficient evidence for reliable recognition. Cross-hospital collaboration could help by allowing the diagnosing institution to use relevant experience stored at other hospitals. However, privacy regulations such as HIPAA, GDPR, and analogous institutional policies restrict the transmission of identifiable clinical text across institutional boundaries~\citep{zhong2025considerations}. This creates the central communication constraint of cross-hospital rare-disease diagnosis: hospitals need to use distributed, case-specific diagnostic evidence, but the clinical records and retrieved text that contain this evidence cannot be shared.

\textbf{Challenge 1: Existing collaboration relies on textual evidence exchange.}
Medical agent systems provide a natural framework for hospital diagnostic reasoning by retrieving medical knowledge, integrating heterogeneous evidence, and performing multi-step reasoning over clinical records~\citep{tang2024medagents,kim2024mdagents,wang2025amanda}. However, these systems typically communicate through natural-language evidence, such as retrieved cases, summaries, or agent messages. This assumption breaks down in cross-hospital rare-disease diagnosis. Because each rare condition may appear only a few times at any single institution, the evidence most relevant to a new query is often distributed across hospital-local case repositories. Yet the retrieved clinical text that carries this evidence cannot leave the institution that stores it. Hospitals therefore need to share case-specific diagnostic signal for the current query without sending the clinical text that encodes it. Figure~\ref{fig:teaser} summarizes the resulting tradeoff: isolated diagnosis keeps records local but cannot use distributed diagnostic experience, while direct text sharing pools evidence but exposes patient-identifiable clinical records. This leaves a key question unanswered: \emph{what should hospital agents exchange during inference when retrieved clinical text cannot be shared?}

\textbf{Challenge 2: Raw latent states can still leak clinical content.}
Latent model representations offer a natural candidate for this missing communication object. Instead of sending natural-language evidence, hospital agents could send internal model states such as hidden states or KV-cache representations, which recent work has shown can support reasoning and model collaboration~\citep{hao2025coconut,zou2025latentmas,du2026interlat,fu2026c2c}. However, replacing text with latent states does not automatically resolve the leakage concern. Embeddings, hidden states, and KV caches can retain substantial information about the original prompt, including retrieved private cases~\citep{morris2023vec2text,dong2025depthfalseprivacy}. Thus, cross-hospital latent communication should be evaluated not only for diagnostic utility, but also for empirical leakage from transmitted latent states.

To address these two challenges, we introduce \method{}, a latent multi-agent communication framework for cross-hospital rare-disease diagnosis. The key idea is that each hospital agent keeps its private clinical records and retrieved cases local, reasons over local evidence, and sends only a compact latent KV block, rather than retrieved text or raw model states, to a host agent for diagnosis. Compared with raw model states, this compact block reduces how much clinical content is reconstructable from the transmitted representation. The same framework extends to settings where hospitals use different LLM backbones by aligning local latent states to the host agent's latent space. Our contributions are summarized as follows.
\begin{itemize}[leftmargin=*,noitemsep,topsep=0pt]
\item \textbf{Formulation.} We formulate cross-hospital rare-disease diagnosis as a clinical-text-restricted latent communication problem, where hospital agents exchange compact latent blocks rather than retrieved clinical text during inference.
\item \textbf{Method.} We instantiate \method{} in two settings: \methodh{} for same-backbone hospital agents through latent KV distillation, and \methodx{} for cross-family hospital agents through latent alignment.
\item \textbf{Benchmark.} We build \benchname{}, a large-scale cross-hospital rare-disease diagnosis benchmark with hospital-level partitions for evaluating inference-time collaboration under text-sharing constraints.
\item \textbf{Evaluation.} We evaluate both diagnostic utility and empirical clinical-content leakage, showing that \method{} improves cross-hospital diagnostic performance while reducing reconstructable clinical content relative to other latent communication baselines.
\end{itemize}

\section{Related Work} \label{sec: related_work}

  \begin{figure*}[t]
      \centering
      \includegraphics[width=\textwidth]{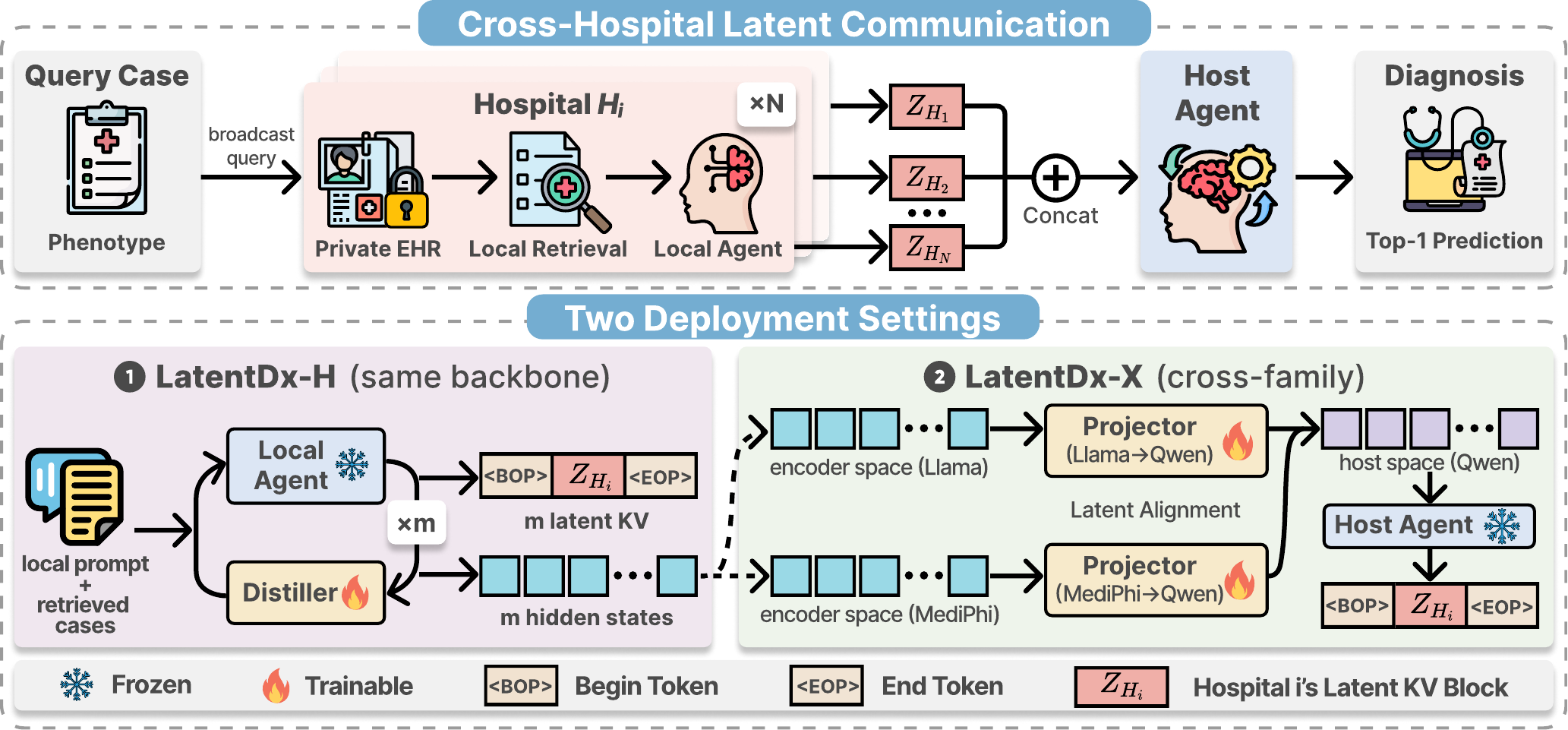}
      \vspace{-0.5cm}
\caption{\textbf{Overview of \method{}.}
\textbf{Top:} Hospital agents keep private EHR records local, retrieve relevant local cases, and send compact latent blocks to a host agent for diagnosis. \textbf{Bottom:} \method{} supports two deployment settings: \methodh{} uses latent KV distillation for same-backbone hospital agents, while \methodx{} uses lightweight projectors for cross-family latent alignment when hospitals use different LLM backbones.}
      \vspace{-0.3cm}
      \label{fig:overview}
  \end{figure*}

\paragraph{Medical diagnostic agents.}
Recent medical LLM agents improve clinical reasoning through retrieval, multi-agent collaboration, and knowledge-grounded decision making~\citep{tang2024medagents,kim2024mdagents,wang2025amanda}. This direction has been extended to evidence-based diagnostic workflows, EHR-grounded decision support, hospital-scale agent simulation, and multi-disciplinary rare-disease reasoning~\citep{wang2026medagentpro,wang2025colacare,li2024agenthospital,chen2026rareagents}. For rare diseases, DeepRare advances agentic differential diagnosis, while PhenoBrain and RareBench provide phenotype-based diagnostic modeling and benchmark evaluation~\citep{zhao2026deeprare,mao2025phenobrain,chen2024rarebench}. These systems also build on phenotype-driven pipelines that match patient phenotypes to candidate diseases or genes~\citep{birgmeier2020amelie,robinson2020lirical,zhai2023phen2disease}. Despite these advances, existing diagnostic agents and phenotype-based systems mainly operate within a single clinical trust boundary or rely on textual clinical evidence. They therefore do not define what hospital agents should exchange during inference when private clinical text cannot be shared across institutions.

\paragraph{Privacy-aware hospital collaboration.}
Privacy-aware collaboration has a long history in medical AI, especially through federated learning and hospital deployment frameworks. Federated LLM methods share gradients, adapters, LoRA updates, or model parameters rather than patient records~\citep{ye2024openfedllm,zhang2024fedit,liu2025dplora}, and medical FL infrastructure has been developed for real cross-institution settings~\citep{pati2022federated,roth2023flare,cremonesi2023fedbiomed}. These methods aggregate distributed experience into training-time updates, but they do not let a diagnosing hospital condition the current query on case-specific evidence retrieved from other hospitals' local databases. A complementary line studies privacy-aware LLM inference by splitting models, adding local privacy noise, or compressing transmitted representations~\citep{du2023dpforward,mai2024splitdenoise,yang2024pfid}. These approaches protect single-client inference or delegated computation, whereas \method{} operates in a different regime: patient-specific cross-hospital collaboration during inference, where hospitals exchange diagnostic signal without transmitting clinical text.

\paragraph{Latent communication.}
Recent work has explored whether LLMs can reason or communicate through continuous internal representations rather than text. Coconut performs continuous reasoning by feeding hidden states back as input embeddings~\citep{hao2025coconut}, while related work studies soft prompts, pause tokens, and compressed KV-style representations~\citep{li2021prefixtuning,mu2023gist,goyal2024pausetokens,zhang2025softthinking,kuzina2026kava}. In multi-agent and cross-model settings, LatentMAS and Interlat exchange latent states, while Cache-to-Cache and KV-alignment methods learn KV-cache projection or fusion across LLMs~\citep{zou2025latentmas,du2026interlat,fu2026c2c,dery2026kvalign}. Further work on systems shows that KV caches can be compressed, reused, and transmitted efficiently~\citep{zhang2023h2o,xiao2024streamingllm,liu2024cachegen, chen2026federated}. Together, these studies establish latent communication as a promising model-collaboration mechanism, but mainly target general reasoning, efficiency, or same-input cross-model communication rather than cross-hospital diagnosis, where agents retrieve private cases and require joint evaluation of diagnostic utility and empirical leakage.
\section{Method}
\label{sec:method}

A central question in cross-hospital rare-disease diagnosis is what hospital agents should exchange during inference. This communication object should avoid both textual evidence exchange and raw prompt-length KV exchange: retrieved clinical text cannot be shared across hospitals, while full KV states can still expose prompt-derived clinical content. \method{} meets these requirements by having each hospital agent keep retrieved cases, local prompts, and prompt-length KV caches local, while sending only a fixed-length diagnosis-oriented latent block instead of retrieved text or raw KV states. We first formalize the inference and reconstruction-leakage settings in Sec.~\ref{sec:problem-setup}, then present \methodh{} for same-backbone latent distillation in Sec.~\ref{sec:method-h} and \methodx{} for cross-family latent alignment in Sec.~\ref{sec:method-x}.

\subsection{Problem Setup}
\label{sec:problem-setup}

\paragraph{Task formulation.}
We consider a cross-hospital network of hospital agents $\mathcal{S}=\{H_1,\ldots,H_N\}$. Each hospital agent $H_i$ has a local LLM backbone and a private case database $\mathcal{D}_i$ of historical patient records. Given a query case, we represent the query only as a standardized HPO phenotype set $q_{\mathrm{HPO}}$, whose codes are deterministically rendered into canonical HPO term names for prompting. The query is therefore not a free-text clinical note. This structured phenotype representation is broadcast to all hospital agents. Each hospital agent then retrieves relevant cases only from its own private database $\mathcal{D}_i$ and constructs a local prompt $x_i$ from the query phenotypes and retrieved local cases. The local prompt $x_i$, raw clinical text, and retrieved patient records never leave the hospital that stores them. During inference, each hospital agent sends only a transmitted representation $\mathcal{T}_{H_i}$ to the host agent, typically the query hospital agent, and the goal is to produce a single OMIM-coded diagnosis:
\begin{equation}
\hat{y}=\mathrm{Host}\!\left(q_{\mathrm{HPO}},\,\bigoplus_{i=1}^{N}\mathcal{T}_{H_i}\right), \;
\hat{y}\in\mathcal{Y}_{\mathrm{OMIM}}.
\end{equation}
Here, $\mathcal{T}_{H_i}$ denotes the transmitted object. In \method{}, $\mathcal{T}_{H_i}=\mathcal{Z}_{H_i}$ is the compact latent block, while in the raw-KV baseline, $\mathcal{T}_{H_i}=\mathcal{M}_{H_i}$ is the prompt-length KV cache.

\paragraph{Leakage setting.}
We evaluate leakage under a passive observer of the inter-hospital communication channel. The observer can read the transmitted object $\mathcal{T}_{H_i}$ from each hospital agent, such as a prompt-length KV cache in the raw-KV baseline or a compact latent block in \method{}. The observer knows the participating LLM architectures and weights, but cannot access private hospital databases. Given $\mathcal{T}_{H_i}$, an attacker $\mathcal{A}$ targets a secret $s_i$ held by hospital $H_i$:
\begin{equation}
\hat{s}_i=\mathcal{A}(\mathcal{T}_{H_i}),
\qquad
\rho_i=\mathrm{score}(\hat{s}_i,s_i).
\end{equation}
Attacks differ in what they take $s_i$ to be: the local prompt content $x_i$ including retrieved clinical evidence, the retrieved disease label alone, or the bit recording whether a given case lies in $\mathcal{D}_i$. Our first instantiation of $\mathcal{A}$ is a frozen-LLM continuation attack, which injects the intercepted state as past key--value states into the frozen LLM that produced them, prompts it to repeat the previous hospital case content, and greedily decodes; here $s_i=x_i$ and $\mathrm{score}$ is token F1 or BERTScore. Sec.~\ref{sec:expt:privacy} adds four stronger attackers, including ones that train a dedicated model on the transmitted representation.

\begin{figure}[t]
      \centering
      \includegraphics[width=\columnwidth]{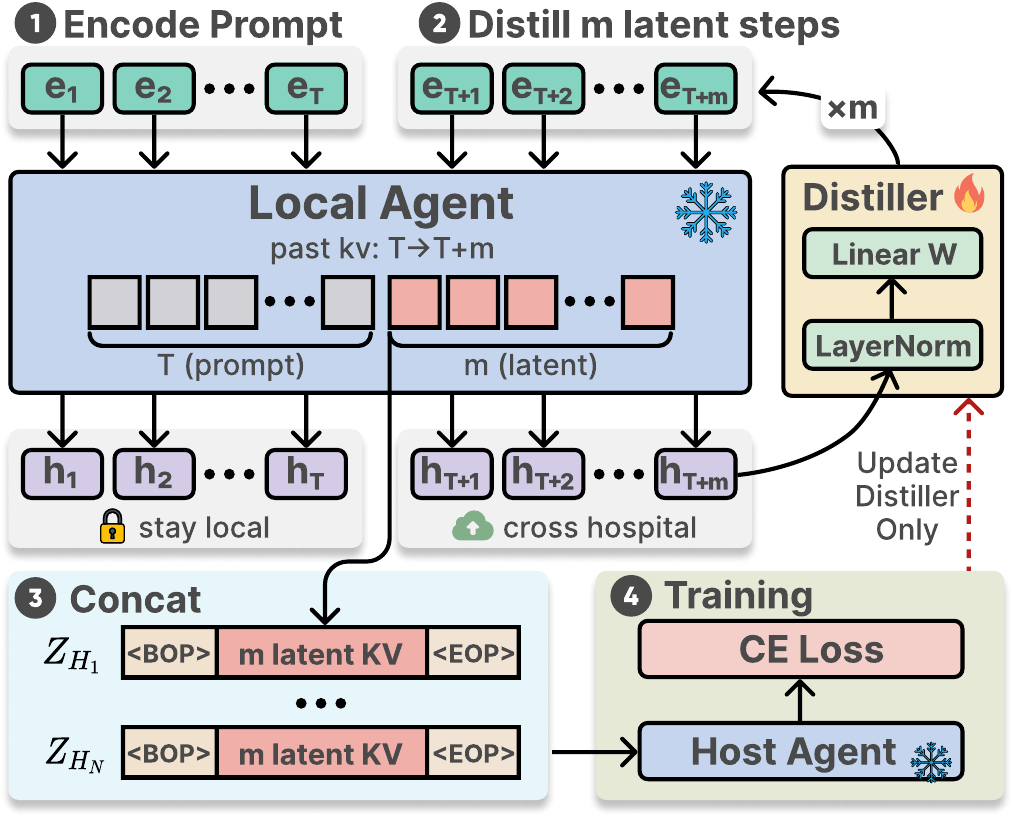}
      \vspace{-0.5cm}
\caption{\small\textbf{\methodh{}: same-backbone latent distillation.}
A hospital agent encodes its local prompt with a frozen LLM backbone, keeping the prompt-length states local. A trainable distiller generates $m$ latent steps, and only the corresponding $m$ latent KV positions are sent as the hospital's compact block. The host agent aggregates these blocks for diagnosis. Training uses diagnosis cross-entropy and updates only the distiller and boundary embeddings.}
\vspace{-0.3cm}
      \label{fig:distill}
  \end{figure}

\subsection{\methodh{}: Same-Backbone Latent Distillation}
\label{sec:method-h}
We first consider the same-backbone setting, where all hospital agents use the same LLM backbone. Their KV spaces are naturally aligned, so the main challenge is compression: each hospital agent must summarize its local prompt state into a compact latent block rather than transmitting the full prompt-length KV cache. Figure~\ref{fig:distill} illustrates this same-backbone distillation pipeline.

\paragraph{Local retrieval and prompt encoding.}
Each hospital agent receives the broadcast phenotype query $q_{\mathrm{HPO}}$ and retrieves phenotypically similar cases only from its private database $\mathcal{D}_i$ using an HPO-based retrieval procedure. The retrieved cases are combined with the query phenotypes to form a local prompt $x_i$, which remains inside hospital $H_i$. A frozen local LLM backbone then encodes $x_i$ and produces a prompt-length KV cache
\begin{equation}
\mathcal{M}_{H_i}
\!=\!
\big\{(\mathbf{K}^{(l)}_{H_i},\mathbf{V}^{(l)}_{H_i})\big\}_{l=1}^{L}, \;
\mathbf{K}^{(l)}_{H_i},\mathbf{V}^{(l)}_{H_i}\in\mathbb{R}^{T_i\times d},
\end{equation}
where $L$ is the number of layers, $T_i$ is the local prompt length, $d$ is the per-layer KV dimension, and $\mathbf{K}^{(l)}_{H_i},\mathbf{V}^{(l)}_{H_i}$ are the layer-$l$ key and value states. Directly transmitting $\mathcal{M}_{H_i}$ gives the raw-KV baseline: it shares no explicit clinical text, but exposes internal representations of the entire local prompt, including retrieved private cases. \methodh{} instead keeps $\mathcal{M}_{H_i}$ local and transmits only a compact latent block $\mathcal{Z}_{H_i}$ distilled from it. Details of the retrieval procedure are provided in Appendix~\ref{app:retrieval}.

\paragraph{Distilling compact latent blocks.}
Within each hospital agent, starting from the final prompt hidden state $h_{T_i}$, we generate compact latent positions autoregressively. We use a minimal distiller so that performance gains can be attributed to the compact latent communication mechanism rather than to additional model capacity. The distiller $\phi$ maps the hidden state back to the input-embedding space,
\begin{equation}
\label{eq:distiller}
\phi(h)=W\cdot\mathrm{LayerNorm}(h),
\end{equation}
where $W\in\mathbb{R}^{d\times d}$ is a bias-free linear projection. The resulting vector is fed into the frozen local LLM backbone as the next input embedding without decoding text. Repeating this procedure for $m$ latent steps produces $m$ latent positions. \methodh{} then keeps only the KV states corresponding to these latent positions, denoted by $\tilde{\mathcal{M}}_{H_i}$. This compact cache has the same layer-wise key--value structure as $\mathcal{M}_{H_i}$, but contains only the $m$ latent positions rather than all $T_i$ prompt positions. For host-side consumption, we wrap it with learned boundary embeddings to form the hospital block
\begin{equation}
\mathcal{Z}_{H_i}
=
\big[\langle\textsc{bop}\rangle,\tilde{\mathcal{M}}_{H_i},\langle\textsc{eop}\rangle\big].
\end{equation}
Here, $\langle\textsc{bop}\rangle$ and $\langle\textsc{eop}\rangle$ are learned begin- and end-of-block embeddings that delimit each hospital agent's contribution in the latent chain. They are shared interface parameters already stored by the host and therefore add no per-query transmission. Each hospital transmits only $\tilde{\mathcal{M}}_{H_i}$, a fixed cache of $m$ latent positions independent of the local prompt length $T_i$. The host agent aggregates $\mathcal{Z}_{H_1}\oplus\cdots\oplus\mathcal{Z}_{H_N}$ and produces the final diagnosis.

\paragraph{Training objective.}
We train the distiller $\phi$ and the boundary embeddings using diagnosis cross-entropy on the host agent's output. Let $y=(y_1,\ldots,y_{|y|})$ denote the target answer sequence. The loss is
\begin{equation}
\label{eq:distill-loss}
\mathcal{L}_{\mathrm{Distill}}
=
-\sum_{t=1}^{|y|}
\log p\big(y_t \mid \mathcal{Z}_{H_{1:N}},\, y_{<t}\big),
\end{equation}
where $\mathcal{Z}_{H_{1:N}}$ denotes the concatenated latent blocks from all hospitals. The loss supervises the final diagnosis tokens generated by the host agent. Since the intermediate latent positions have no target text labels, they are learned indirectly through gradients from this diagnosis loss. During training, we update only the distiller and boundary embeddings, while all local and host LLM backbones remain frozen. We train the latent interface centrally on the public training split under simulated hospital partitions, with train and test cases disjoint.

\subsection{\methodx{}: Cross-Family Latent Alignment}
\label{sec:method-x}
In practice, hospitals may deploy different LLM backbones due to local infrastructure or domain adaptation. Different backbones may have different hidden dimensions and representation statistics, so a compact latent block produced by one hospital agent may not be directly consumable by the host agent. \methodx{} addresses this mismatch through cross-family latent alignment.

\paragraph{Cross-family projector.} To handle cross-family settings, \methodx{} uses a lightweight projector to map compact latent states into the host agent's input-embedding space. Let $\mathcal{E}$ denote an encoder backbone family and $\mathcal{H}$ denote the host backbone family. Let $\mathbf{h}^{(\mathcal{E})}_{1:m_{\mathcal{E}}}\in\mathbb{R}^{m_{\mathcal{E}}\times d_{\mathcal{E}}}$ denote the concatenated compact latent states from hospital agents using encoder family $\mathcal{E}$, where $m_{\mathcal{E}}$ is the total number of latent positions in that group. The projector maps it to host input embeddings
\begin{equation}
\mathbf{e}^{(\mathcal{H})}_{1:m_{\mathcal{E}}}
=
\psi_{\mathcal{E}\to\mathcal{H}}\big(\mathbf{h}^{(\mathcal{E})}_{1:m_{\mathcal{E}}}\big)
\in
\mathbb{R}^{m_{\mathcal{E}}\times d_{\mathcal{H}}}.
\end{equation}
In our implementation, $\psi_{\mathcal{E}\to\mathcal{H}}$ is a token-wise LayerNorm followed by a bias-free linear projection. The projected sequence is wrapped with boundary tokens and passed to the frozen host LLM through input embeddings. The host backbone then materializes host-space latent KV states for final aggregation and diagnosis.

\paragraph{Joint projector training.}
For each host backbone family $\mathcal{H}$, we train one projector $\psi_{\mathcal{E}\to\mathcal{H}}$ for each non-host encoder family $\mathcal{E}$. During training, each batch samples a hospital-family composition from the available backbone families, while ensuring that at least one non-host family is present. Native-family hospital agents use the host family's frozen latent distiller, while non-host-family agents use their frozen encoder-side latent distillers followed by the corresponding cross-family projector. The host agent aggregates all native and projected latent blocks and is supervised with the same diagnosis cross-entropy objective as Eq.~\ref{eq:distill-loss}. Only the projectors and boundary embeddings are updated. The encoder LLM backbones, encoder-side distillers, and host LLM backbone remain frozen. The resulting projectors can be reused across different hospital compositions at inference time, without retraining for each mixture.


\begin{table*}[t!]
\centering
\small
\setlength{\tabcolsep}{4.5pt}
\renewcommand{\arraystretch}{1}
\caption{\textbf{Diagnostic utility on \benchname{}.}
Accuracy (Acc), macro F1 (F1) and LLM-as-judge score (Judge) on the three source cohorts and on the full test set (Overall). \hlfirst{Best} and \hlsecond{second best} method per column, ranked separately for each backbone; tied methods are both marked. \llmname{MediPhi} results are in Table~\ref{app:full-utility}.}

\label{tab:utility-main}
\vspace{-3pt}
\resizebox{\textwidth}{!}{%
\begin{tabular}{cl|ccc|ccc|ccc!{\vrule width 0.8pt}ccc}
\Xhline{1.2pt}
\rowcolor{headcolor}
  & & \multicolumn{3}{c|}{\hgroup{RareBench}}
  & \multicolumn{3}{c|}{\hgroup{Zenodo}}
  & \multicolumn{3}{c!{\vrule width 0.8pt}}{\hgroup{Phenopacket}}
  & \multicolumn{3}{c}{\hgroup{Overall}} \\
\rowcolor{headcolor}
\hgroup{Model} & \quad\hgroup{Method} & \hcol{Acc} & \hcol{F1} & \hcol{Judge} & \hcol{Acc} & \hcol{F1} & \hcol{Judge} & \hcol{Acc} & \hcol{F1} & \hcol{Judge} & \hcol{Acc} & \hcol{F1} & \hcol{Judge} \\
\Xhline{1pt}
\multirow{12}{*}{\mdl{Qwen3}{4B}}
 & \multicolumn{13}{c}{\cellcolor{catcolor}\textit{Non-collaborative}} \\
  & \quad \mname{Zero-shot} & 0.11 & 0.08 & 0.30 & 0.03 & 0.05 & 0.37 & 0.01 & 0.01 & 0.18 & 0.03 & 0.02 & 0.24 \\
  & \quad \mname{Single RAG} & 0.53 & 0.37 & 0.58 & 0.55 & 0.44 & 0.59 & 0.58 & 0.51 & 0.70 & 0.57 & 0.47 & 0.66 \\
 & \multicolumn{13}{c}{\cellcolor{catcolor}\textit{Text / structured sharing}} \\
  & \quad \mname{Dist.\ RAG} & 0.56 & 0.44 & \hlsecond{0.72} & 0.65 & 0.62 & \hlsecond{0.71} & 0.61 & 0.54 & \hlsecond{0.72} & 0.61 & \hlsecond{0.53} & \hlfirst{0.72} \\
  & \quad \mname{Global RAG} & 0.58 & 0.41 & 0.68 & 0.63 & 0.61 & 0.67 & \hlsecond{0.62} & 0.55 & \hlsecond{0.72} & \hlsecond{0.62} & \hlsecond{0.53} & 0.70 \\
  & \quad \mname{Struct.\ RAG} & 0.61 & 0.45 & 0.70 & 0.64 & 0.56 & 0.66 & \hlfirst{0.64} & \hlsecond{0.56} & \hlfirst{0.73} & \hlfirst{0.64} & 0.52 & \hlsecond{0.71} \\
  & \quad \mname{TextMAS} & 0.53 & 0.37 & 0.61 & 0.49 & 0.49 & 0.65 & 0.56 & 0.48 & \hlsecond{0.72} & 0.54 & 0.44 & 0.69 \\
 & \multicolumn{13}{c}{\cellcolor{catcolor}\textit{Latent communication}} \\
  & \quad \mname{Raw KV} & \hlsecond{0.63} & 0.45 & \hlfirst{0.75} & \hlfirst{0.67} & \hlfirst{0.64} & \hlfirst{0.73} & 0.56 & 0.49 & 0.70 & 0.59 & 0.48 & \hlfirst{0.72} \\
  & \quad \mname{LatentMAS} & 0.61 & \hlsecond{0.46} & \hlsecond{0.72} & \hlsecond{0.66} & \hlsecond{0.63} & \hlsecond{0.71} & 0.57 & 0.51 & 0.71 & 0.60 & 0.49 & \hlsecond{0.71} \\
  & \quad \method{} & \hlfirst{0.72} & \hlfirst{0.61} & \hlfirst{0.75} & 0.59 & 0.61 & 0.67 & \hlfirst{0.64} & \hlfirst{0.63} & \hlfirst{0.73} & \hlfirst{0.64} & \hlfirst{0.61} & \hlfirst{0.72} \\
\Xhline{1pt}
\multirow{12}{*}{\mdl{Llama-3.2}{3B}}
 & \multicolumn{13}{c}{\cellcolor{catcolor}\textit{Non-collaborative}} \\
  & \quad \mname{Zero-shot} & 0.02 & 0.01 & 0.21 & 0.02 & 0.02 & 0.09 & 0.00 & 0.00 & 0.10 & 0.01 & 0.00 & 0.11 \\
  & \quad \mname{Single RAG} & 0.37 & 0.36 & 0.53 & 0.28 & 0.29 & 0.43 & 0.43 & 0.37 & 0.62 & 0.39 & 0.34 & 0.57 \\
 & \multicolumn{13}{c}{\cellcolor{catcolor}\textit{Text / structured sharing}} \\
  & \quad \mname{Dist.\ RAG} & 0.44 & 0.40 & 0.60 & \hlsecond{0.37} & 0.44 & 0.62 & 0.47 & 0.40 & 0.64 & 0.45 & 0.40 & 0.63 \\
  & \quad \mname{Global RAG} & 0.46 & \hlsecond{0.41} & 0.53 & 0.36 & \hlsecond{0.45} & 0.56 & \hlsecond{0.50} & \hlsecond{0.44} & 0.66 & \hlsecond{0.47} & \hlsecond{0.42} & 0.62 \\
  & \quad \mname{Struct.\ RAG} & \hlsecond{0.53} & 0.35 & \hlsecond{0.63} & 0.31 & 0.35 & \hlsecond{0.65} & 0.43 & 0.37 & \hlsecond{0.69} & 0.42 & 0.34 & \hlsecond{0.67} \\
  & \quad \mname{TextMAS} & 0.33 & 0.24 & 0.49 & 0.33 & 0.39 & 0.48 & 0.31 & 0.25 & 0.47 & 0.32 & 0.25 & 0.47 \\
 & \multicolumn{13}{c}{\cellcolor{catcolor}\textit{Latent communication}} \\
  & \quad \mname{Raw KV} & 0.28 & 0.24 & 0.46 & 0.27 & 0.39 & 0.50 & 0.40 & 0.35 & 0.64 & 0.35 & 0.32 & 0.58 \\
  & \quad \mname{LatentMAS} & 0.23 & 0.20 & 0.44 & 0.26 & 0.34 & 0.60 & 0.38 & 0.35 & 0.62 & 0.33 & 0.31 & 0.59 \\
  & \quad \method{} & \hlfirst{0.75} & \hlfirst{0.64} & \hlfirst{0.77} & \hlfirst{0.70} & \hlfirst{0.62} & \hlfirst{0.74} & \hlfirst{0.73} & \hlfirst{0.72} & \hlfirst{0.78} & \hlfirst{0.73} & \hlfirst{0.68} & \hlfirst{0.77} \\
\Xhline{1.2pt}
\end{tabular}}
\vspace{-0.3cm}
\end{table*}

\section{Experiments}
\label{sec:exp}

The experiments below are organized around one question: whether \method{} retains the diagnostic benefit while limiting what an adversary recovers from the transmitted state.

\subsection{Benchmark Construction}
\label{sec:benchmark}

We build \benchname{} from three public rare-disease case repositories: RareBench~\citep{chen2024rarebench}, the Zenodo phenotype--disease collection~\citep{chimirri2025consistent}, and the GA4GH Phenopacket Store~\citep{danis2025corpus}. Each case carries HPO codes and a ground-truth OMIM identifier; after label resolution, deduplication and frequency filtering, $8{,}022$ cases across $235$ OMIM diseases remain, split $90/5/5$ into training, validation and test. We partition the training cases into five hospital-local databases, and at inference each agent retrieves only from its own $\mathcal{D}_i$, so diagnostic evidence is distributed across hospitals while retrieved clinical text stays local. The main results use a balanced partition. Appendix~\ref{app:skew} repeats them under a non-IID Dirichlet($\alpha{=}0.3$) partition in which each disease concentrates in a few hospitals, raising the mean per-disease single-hospital share from $0.20$ to $0.65$; \method{} keeps the strongest overall utility on both hosts there as well.

\benchname{} is a simplified setting: each case arrives as a list of canonical HPO term names. Real hospital evidence is prose, but the public rare-disease repositories are phenotype-coded rather than narrative, so we build \benchnametext{} to study that setting. For every diagnosis instance, both the query and the hospital-retrieved records are rendered as clinical narratives. Cohorts, hospital partitions, and retrieval budget remain fixed, so only the surface form of the phenotype evidence changes. Construction details for both corpora are in Appendices~\ref{app:benchmark} and~\ref{app:note-corpus}.

\subsection{Experimental Setup}
\label{sec:expt:setup}

\paragraph{Baselines.}
We compare \method{} against eight baselines in three groups, all defined in Appendix~\ref{app:baseline-defs}. \ding{182}~\textbf{Non-collaborative} methods see only the query hospital: \mname{Zero-shot} and \mname{Single RAG}. \ding{183}~\textbf{Text / structured sharing} methods send retrieved case text, structured retrieval outputs or natural-language assessments across hospital boundaries: \mname{Dist.\ RAG}, \mname{Struct.\ RAG}, \mname{TextMAS}, and \mname{Global RAG}, which pools every hospital database into a single store and therefore acts as a centralized upper bound. \ding{184}~\textbf{Latent communication} methods send latent state: \mname{Raw KV} and \mname{LatentMAS}. Every method runs on \llmname{Llama-3.2-3B}, \llmname{Qwen3-4B} and \llmname{MediPhi}.

\paragraph{Implementation details.}
We set $m{=}32$ latent positions in all main experiments and use $N{=}3$ hospital agents unless otherwise specified, with the total retrieval budget fixed across all retrieval-based methods so that the comparison isolates the communication mechanism. We report diagnostic utility as accuracy, macro F1 and an LLM-as-judge score, and leakage in three kinds, one per secret the attacker targets. Appendix~\ref{app:metrics} defines every metric and Appendix~\ref{app:impl} gives the full configuration.

\begin{figure}[t!]
\centering
\includegraphics[width=\columnwidth]{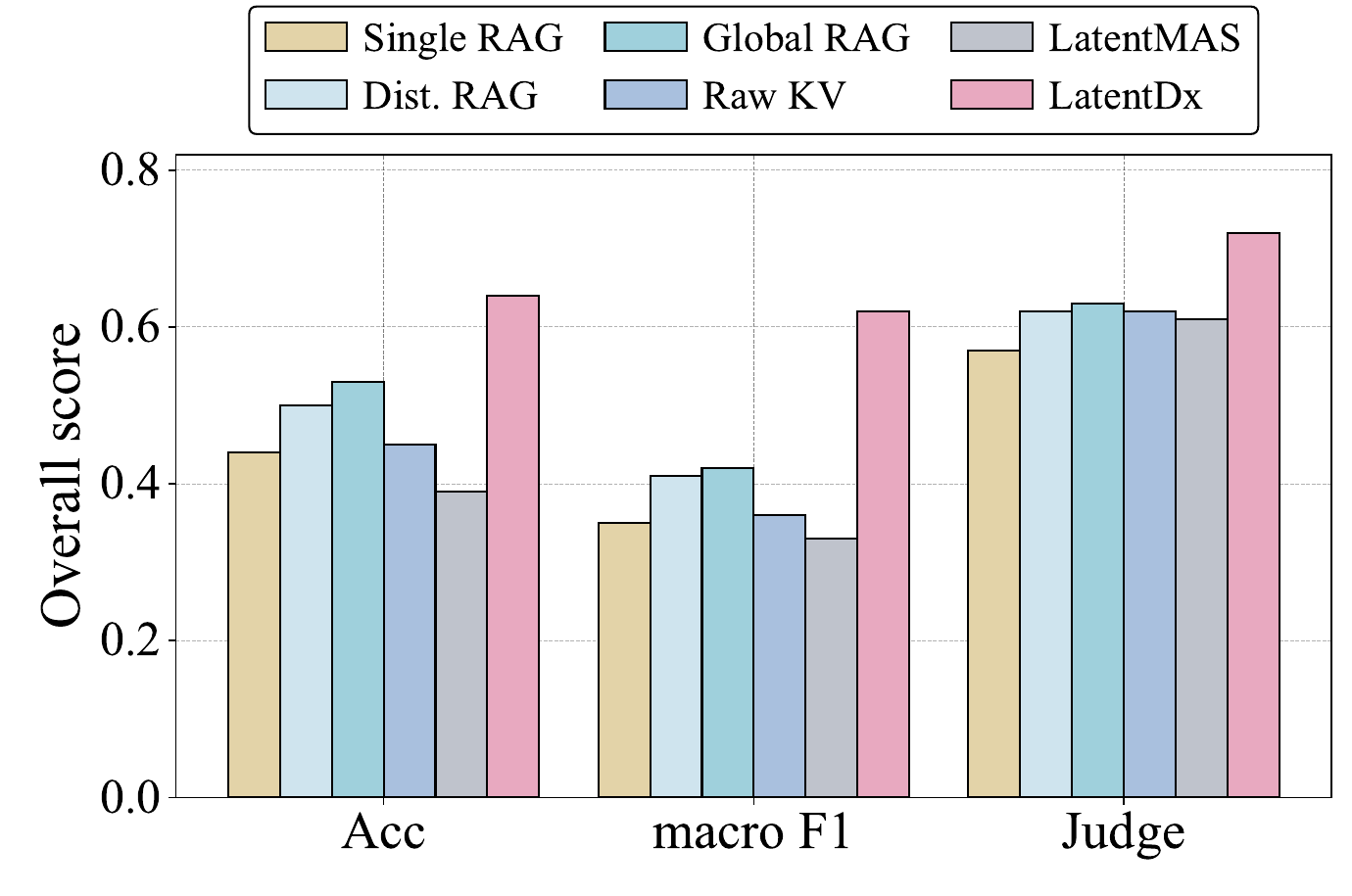}
\vspace{-0.7cm}
\caption{\textbf{\benchnametext{}.} Overall scores on \llmname{Llama-3.2} with both the query and the hospital-retrieved records rendered as clinical narratives. All three hosts in Appendix~\ref{app:note-corpus}.}
\label{fig:freetext}
\vspace{-0.5cm}
\end{figure}

\subsection{Diagnostic Utility}
\label{sec:expt:utility}

Table~\ref{tab:utility-main} reports diagnostic performance in the same-backbone setting, where all hospital agents and the host use the same LLM backbone. Across all three host LLMs, including \llmname{MediPhi} results in Appendix~\ref{app:utility}, \method{} achieves the strongest overall utility, with the best overall accuracy and macro F1 and competitive or best judge scores.

The size of the gain varies across backbones and cohorts, but the aggregate pattern does not. Per-cohort scores divide the $401$-case test set three ways, so individual cells carry wide intervals: on \llmname{Qwen3}, \method{} trails \mname{Raw KV} on the Zenodo cohort while still leading on overall accuracy and macro F1, and the paired bootstrap in Appendix~\ref{app:bootstrap-ci} confirms the macro F1 gain with accuracy and judge score statistically comparable to the strongest baseline. The narrower \llmname{Qwen3} margin tracks a less stable distillation: under the same training configuration, its distillation loss oscillates more than that of \llmname{Llama-3.2}. We therefore read Table~\ref{tab:utility-main} as one half of a joint utility--leakage profile rather than a demand for uniform dominance.

\begin{table}[t!]
\centering
\small
\setlength{\tabcolsep}{4.5pt}
\renewcommand{\arraystretch}{1.18}
\caption{\textbf{Effect of cross-family heterogeneity.}
We vary the model-family composition of three hospital agents. Deg0, Deg1, and Deg2 use one, two, and three model families, respectively. $\Delta$Acc reports the absolute accuracy change from Deg0 on the percentage-point scale. Q = \llmname{Qwen3-4B}, L = \llmname{Llama-3.2-3B}, and M = \llmname{MediPhi}.}
\label{tab:heterogeneity}
\vspace{-0.1cm}
\resizebox{\columnwidth}{!}{%
\begin{tabular}{ll|ccc|c}
\Xhline{1.2pt}
\rowcolor{headcolor}
\hgroup{Host} & \hgroup{Degree (comp.)} & \hcol{Acc} & \hcol{F1} & \hcol{Judge} & \hgroup{$\Delta$Acc} \\
\Xhline{1pt}
\multirow{3}{*}{\llmname{Qwen3}}
 & Deg0 (Q:Q:Q)   & 0.64 & 0.61 & 0.72 & \cellcolor{catcolor}-- \\
 & Deg1 (Q:L:Q)   & 0.63 & 0.60 & 0.70 & \cellcolor{catcolor}$-0.5$ \\
 & Deg2 (Q:L:M)   & 0.62 & 0.60 & 0.69 & \cellcolor{catcolor}$-2.2$ \\
\Xhline{0.6pt}
\multirow{3}{*}{\llmname{Llama-3.2}}
 & Deg0 (L:L:L)   & 0.73 & 0.68 & 0.77 & \cellcolor{catcolor}-- \\
 & Deg1 (L:Q:L)   & 0.72 & 0.66 & 0.76 & \cellcolor{catcolor}$-0.7$ \\
 & Deg2 (L:Q:M)   & 0.71 & 0.65 & 0.76 & \cellcolor{catcolor}$-1.6$ \\
\Xhline{1.2pt}
\end{tabular}}
\vspace{-0.3cm}
\end{table}

\begin{table*}[t!]
\centering
\small
\setlength{\tabcolsep}{4.5pt}
\renewcommand{\arraystretch}{1.18}
\caption{\textbf{Leakage under attacks of increasing attacker capability.}
Lower is better. No~tx. denotes an attack-specific no-transmission control: the attacker receives no transmitted representation. It serves as the control baseline for measuring additional leakage attributable to transmission. Learned-attack entries are means over three random seeds. Reduction is the relative drop from \mname{Raw KV} to \method{}. Metrics are not comparable across rows. \llmname{MediPhi} and full details appear in Appendix~\ref{app:leakage-attacks}.}
\label{tab:privacy-main}
\vspace{-0.1cm}
\resizebox{\textwidth}{!}{%
\begin{tabular}{ll|ccccc!{\vrule width 0.8pt}ccccc}
\Xhline{1.2pt}
\rowcolor{headcolor}
 & & \multicolumn{5}{c!{\vrule width 0.8pt}}{\hgroup{\llmname{Qwen3-4B}}}
   & \multicolumn{5}{c}{\hgroup{\llmname{Llama-3.2-3B}}} \\
\rowcolor{headcolor}
\hgroup{Attack} & \hgroup{Metric}
 & \hcol{No tx.} & \hcol{\mname{Raw KV}} & \hcol{\mname{LatentMAS}} & \hcol{\method{}} & \hgroup{Reduction}
 & \hcol{No tx.} & \hcol{\mname{Raw KV}} & \hcol{\mname{LatentMAS}} & \hcol{\method{}} & \hgroup{Reduction} \\
\Xhline{1pt}
\multicolumn{12}{c}{\cellcolor{catcolor}\textit{Training-free attacks}} \\
Prompt reconstruction & Token F1 $\downarrow$
 & 0.05 & 0.60 & 0.60 & 0.07 & \reduc{$\downarrow$88\%}
 & 0.07 & 0.68 & 0.82 & 0.07 & \reduc{$\downarrow$90\%} \\
Template-aware extraction & Disease recall $\downarrow$
 & 0.00 & 0.89 & 0.90 & 0.20 & \reduc{$\downarrow$78\%}
 & 0.00 & 0.88 & 0.75 & 0.16 & \reduc{$\downarrow$82\%} \\
\Xhline{0.6pt}
\multicolumn{12}{c}{\cellcolor{catcolor}\textit{Learned attacks}} \\
Attribute probe & OMIM exact $\downarrow$
 & 0.17 & 0.62 & 0.62 & 0.23 & \reduc{$\downarrow$63\%}
 & 0.17 & 0.61 & 0.60 & 0.20 & \reduc{$\downarrow$67\%} \\
Inversion decoder & OMIM exact $\downarrow$
 & 0.27 & 0.97 & 0.96 & 0.40 & \reduc{$\downarrow$59\%}
 & 0.27 & 0.99 & 0.99 & 0.38 & \reduc{$\downarrow$62\%} \\
Membership inference & TPR@1\%FPR $\downarrow$
 & 0.00 & 0.99 & 0.93 & 0.17 & \reduc{$\downarrow$83\%}
 & 0.00 & 1.00 & 0.83 & 0.03 & \reduc{$\downarrow$97\%} \\
\Xhline{1.2pt}
\end{tabular}}
\end{table*}

\begin{figure*}[t!]
    \centering
    \begin{subfigure}[t]{0.245\textwidth}
        \centering\includegraphics[width=\linewidth]{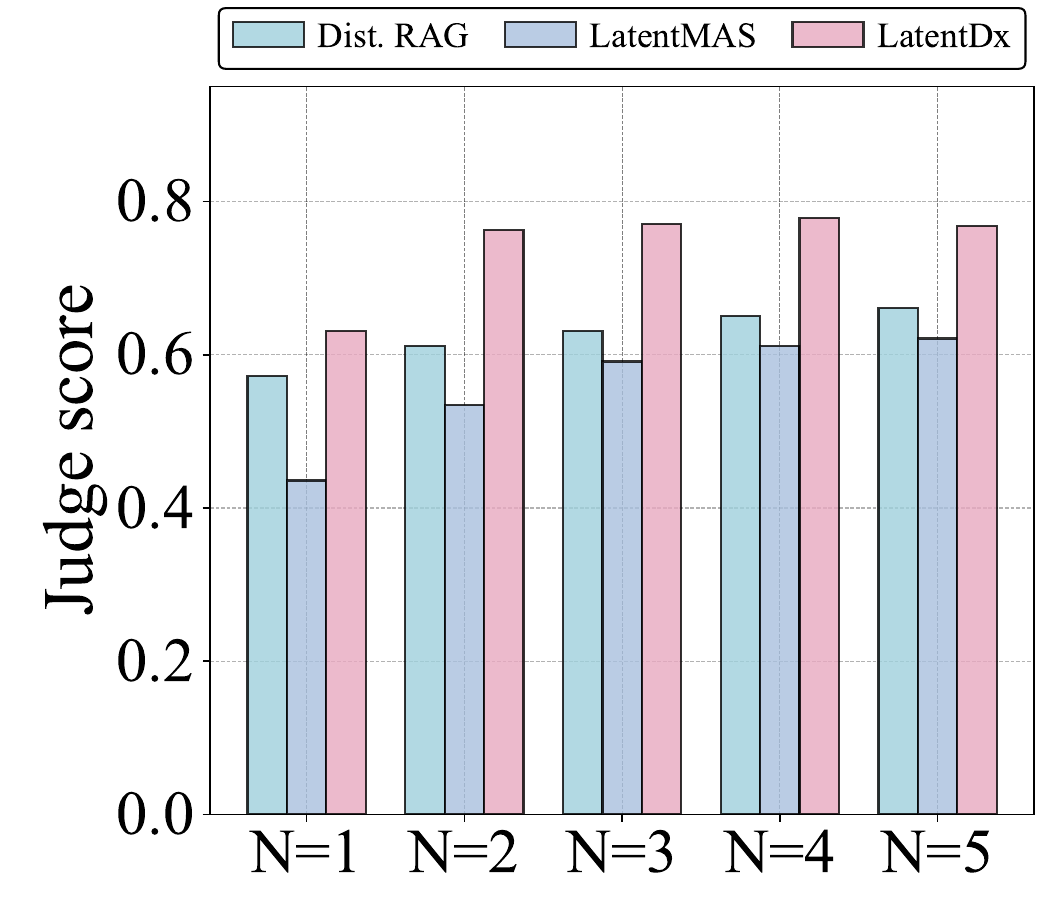}
        \caption{Hospital agents $N$}\label{fig:analysis_n}
    \end{subfigure}\hfill
    \begin{subfigure}[t]{0.245\textwidth}
        \centering\includegraphics[width=\linewidth]{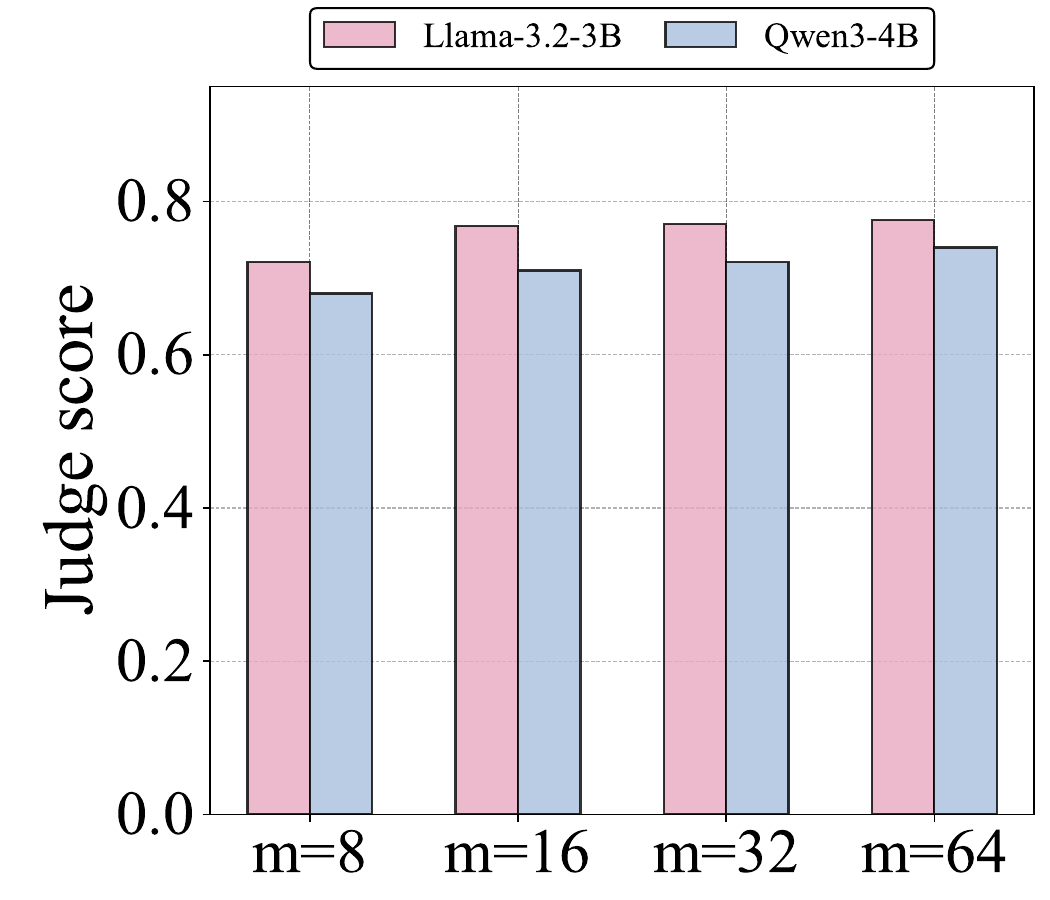}
        \caption{Latent block length $m$}\label{fig:analysis_m}
    \end{subfigure}\hfill
    \begin{subfigure}[t]{0.245\textwidth}
        \centering\includegraphics[width=\linewidth]{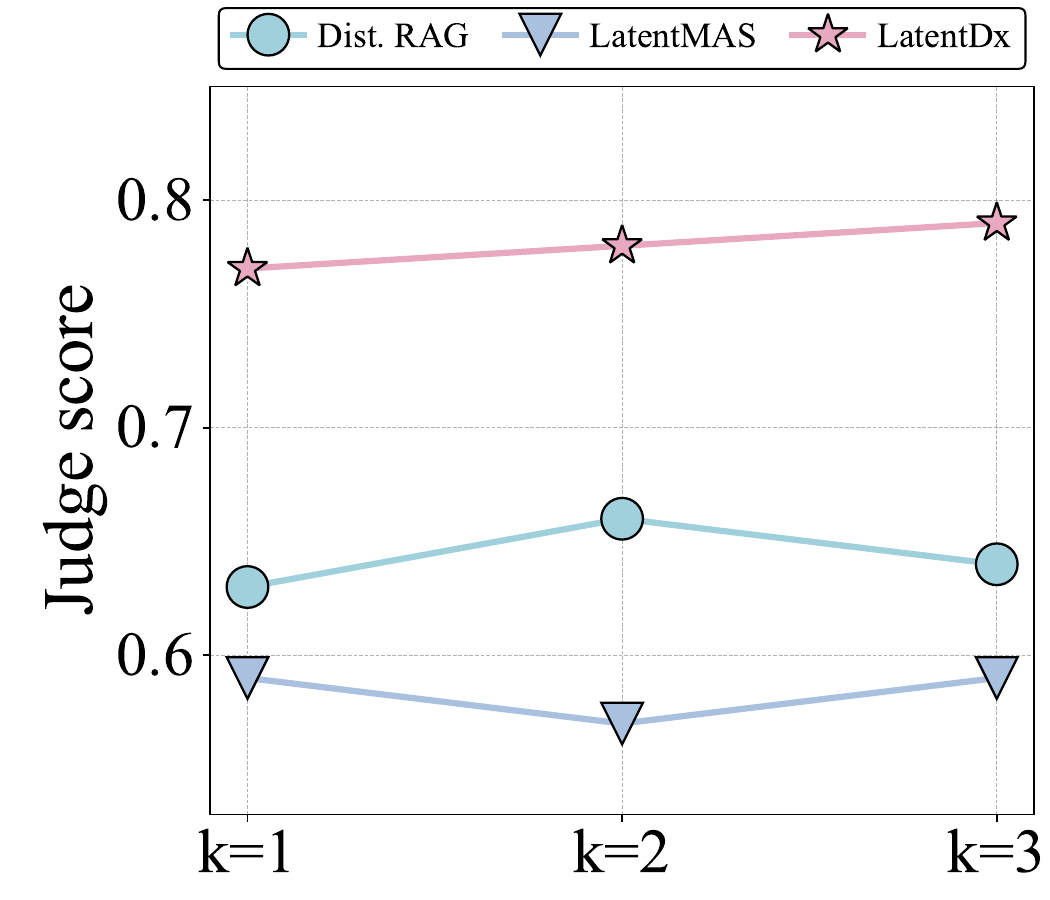}
        \caption{Retrieval depth $k$}\label{fig:analysis_k}
    \end{subfigure}\hfill
    \begin{subfigure}[t]{0.245\textwidth}
        \centering\includegraphics[width=\linewidth]{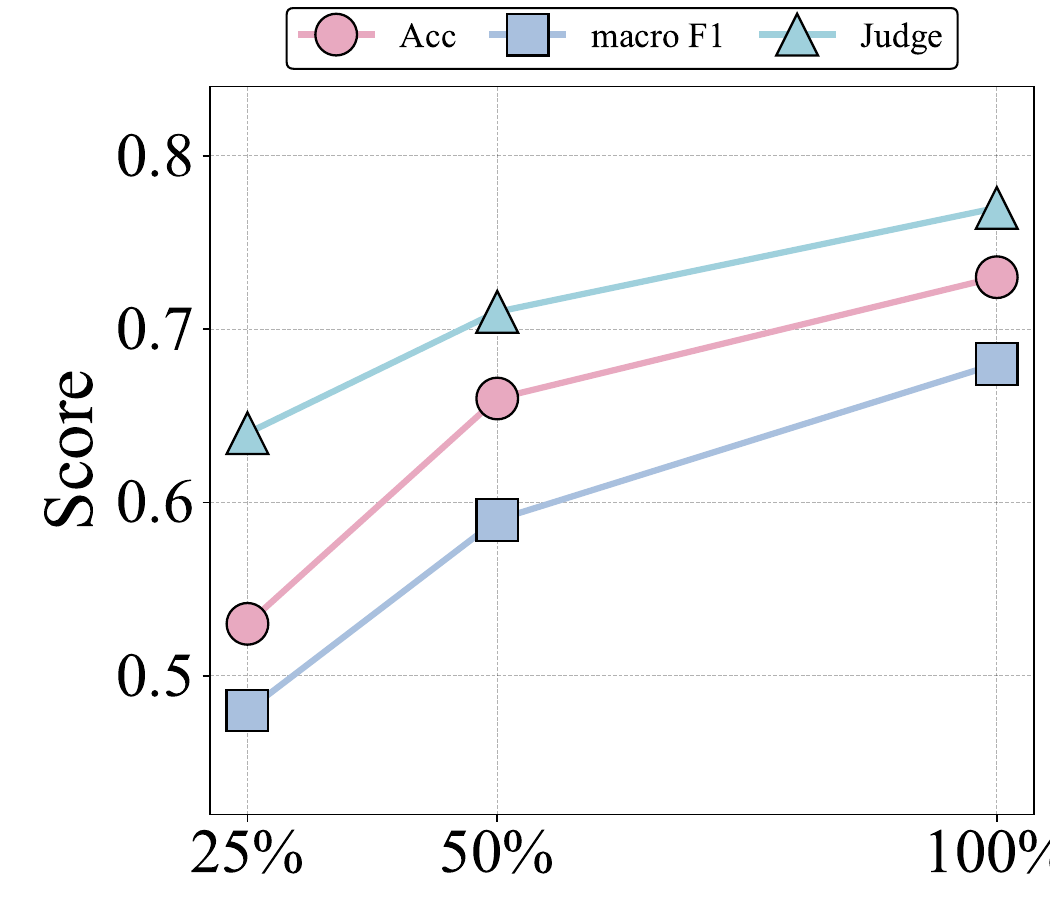}
        \caption{Training data size}\label{fig:analysis_train}
    \end{subfigure}
    \vspace{-0.2cm}
\caption{\textbf{Hyperparameter analysis.} Each panel varies one setting and holds the rest at their main-experiment values. (a) and (c) vary the number of hospital agents $N$ and the retrieval depth $k$ on the \llmname{Llama-3.2} host, comparing \mname{Dist.\ RAG}, \mname{LatentMAS} and \method{}. (b) varies the latent block length $m$ for \method{} on both \llmname{Llama-3.2} and \llmname{Qwen3}. (d) varies the fraction of the training split used to fit \method{}'s latent interface on \llmname{Llama-3.2}, reporting all three utility metrics. Panels (a) to (c) report judge score. Full numbers in Appendix~\ref{app:ablations}.}
    \label{fig:analysis}
    \vspace{-0.3cm}
\end{figure*}

\subsection{Cross-Family Deployment}
\label{sec:expt:crossfamily}

Table~\ref{tab:heterogeneity} evaluates \methodx{} when hospital agents use different LLM backbone families. Performance drops only mildly on both hosts as the composition grows more heterogeneous, so the cross-family projectors retain most diagnosis-relevant signal when mapping non-host latent states into the host space, and they can be reused across hospital compositions without retraining for each mixture.

\subsection{Leakage under Attack}
\label{sec:expt:privacy}

Table~\ref{tab:privacy-main} reports leakage under a ladder of attacks of increasing attacker capability: two training-free attacks that reuse the frozen host LLM and train nothing, and three that train a dedicated model on the transmitted representation. Text-sharing baselines are not listed, since they transmit retrieved clinical text and a passive observer reads the source content directly. Both raw-latent baselines leak under every attack, so operating in latent space is not by itself protective. Reading each cell as the fraction of \mname{Raw KV}'s leakage that survives above the no-transmission floor, \mname{LatentMAS} recovers $83$ to $123\%$, exceeding the raw cache itself on several backbone and attack pairs. This follows from what it transmits: \mname{LatentMAS} runs its own latent steps on top of the prompt KV cache rather than in place of it, so the transmitted object carries the raw cache together with the further state those steps derive from it. Together with Table~\ref{tab:utility-main}, these results show that \method{} preserves diagnostic utility while reducing the clinical content recoverable from transmitted latent states.

\subsection{Hyperparameter Analysis}
\label{sec:expt:analysis}

Figure~\ref{fig:analysis} ablates four settings of the experimental configuration, each varied in isolation on the \llmname{Llama-3.2} host. \ding{182}~\textbf{Hospital agents $N$.} \method{} receives evidence from more hospital-local databases as $N$ grows and improves throughout, staying ahead of every baseline, so compact latent blocks aggregate additional hospital evidence rather than being overwhelmed by more inputs. \ding{183}~\textbf{Latent block length $m$.} Larger blocks carry more diagnosis-relevant information and generally help, but the gains become moderate after $m{=}32$; since longer blocks also lengthen the transmitted state and raise inference cost, we use $m{=}32$ in the main experiments, which Appendix~\ref{app:comm-cost} shows gives a $5$--$6\times$ communication reduction over \mname{Raw KV}. Appendix~\ref{app:system-profile} profiles the rest of the cost: at three hospitals \method{} adds $0.01$s per query and $0.7$GB of peak memory over \mname{Raw KV}, and fitting the interface is a one-time $5.5$ GPU-hours. \ding{184}~\textbf{Retrieval depth $k$.} The text-sharing baselines improve as $k$ grows, since they consume retrieved cases as prompt context, whereas \method{} is flat: it compresses whatever the hospital retrieves into a block of fixed length, so depth changes what is summarised rather than how much is transmitted. \ding{185}~\textbf{Training data.} Performance rises across the whole sweep and has not saturated at the full split, and a quarter of the data already suffices to pass every comparison method. \ding{186}~\textbf{Extended ablations.} Appendix~\ref{app:component-ablation} varies the interface itself and finds that the distiller can stay simple, since a two-layer MLP gains little over LayerNorm followed by one linear projection, while on the host side concatenating the hospital blocks clearly beats mean-pooling them into a single vector. Appendix~\ref{app:trained-bottleneck} then shows what the distiller cannot give up: replacing autoregressive distillation with a one-shot mean-pooled bottleneck, at the same latent budget, collapses accuracy.

\section{Conclusion}
\label{sec:conclusion}

We introduced \method{}, a latent multi-agent communication framework for cross-hospital rare-disease diagnosis. At its core, hospital agents keep retrieved cases and prompt-length KV caches local and exchange only a fixed-length, diagnosis-oriented latent block. Latent space is not by itself protective: both raw-latent baselines leak under all five attacks we run, and one leaks more than the raw KV cache it was meant to improve on. What limits leakage is compression under a diagnosis objective. Under it, \method{} attains the strongest overall utility on every backbone while admitting only a small share of the raw cache's leakage above a no-transmission control, and holds that profile on prose evidence and under skewed disease prevalence. Compact latent blocks therefore look like a promising communication object for collaboration that cannot move clinical text.

\section{Limitations}
\label{sec:limitations}

Our experiments run entirely on public rare-disease resources. We assembled as many as we could find, but three repositories of curated cases do not reproduce the diversity of real clinical records, and the hospitals in our setting are partitions of that pooled corpus rather than institutions with their own referral patterns, coding conventions and documentation styles. The evaluation is therefore a simulation of cross-hospital collaboration over public data. The method itself does not depend on that: a hospital agent encodes whatever its own database returns, so swapping a simulated partition for a real institutional database changes the retrieval index and leaves the latent interface untouched. Validating the setting on real deployments is the natural next step once such access becomes possible.

\benchnametext{} sits under the same constraint. Real hospital evidence is prose, but no public rare-disease corpus supplies narrative cases at this scale, so we rendered the phenotype-coded cases as clinical narratives ourselves. Holding everything but the surface form fixed is what lets us isolate its effect, though generated narratives are cleaner than real notes, carrying no abbreviations, no negation spread across sections, no transcription noise and no clinician-specific phrasing. The corpus therefore measures what prose input costs under controlled conditions, and extending the study to real notes needs only a narrative source, since the pipeline consumes text either way.

\section{Societal Impact}
\label{sec:societal-impact}

\method{} is a research framework intended to assist, rather than replace, clinical judgment. Because rare-disease predictions can be wrong or perform unevenly across diseases, institutions and patient groups, real use would require clinical validation and expert review.

\bibliography{main}

\clearpage
\newpage
\appendix
\section*{Appendix Roadmap}
{\small
\begin{itemize}[leftmargin=*,itemsep=1pt,parsep=0pt,topsep=2pt,partopsep=0pt]
  \item \hyperref[app:data-task]{\textbf{Appendix~\ref*{app:data-task}: Data and task construction}} describes the benchmark sources, filtering and splits, hospital partitions, and local retrieval procedure.
  \item \hyperref[app:setup]{\textbf{Appendix~\ref*{app:setup}: Experimental setup}} defines the baselines and access patterns, gives the prompts and evaluation metrics, and reports the model and training configurations.
  \item \hyperref[app:system]{\textbf{Appendix~\ref*{app:system}: System efficiency and deployment}} reports latency, memory, hospital scaling, interface-training cost, communication volume, and cross-family deployment details.
  \item \hyperref[app:utility-robustness]{\textbf{Appendix~\ref*{app:utility-robustness}: Utility and robustness}} provides full-backbone utility results, free-text evaluation, non-IID hospital experiments, and bootstrap confidence intervals.
  \item \hyperref[app:analysis]{\textbf{Appendix~\ref*{app:analysis}: Representation analysis and ablations}} studies the trained bottleneck, retrieval and data sensitivity, component design, and latent-space structure.
  \item \hyperref[app:privacy]{\textbf{Appendix~\ref*{app:privacy}: Privacy and leakage}} defines the attack settings and reports reconstruction, attribute, probing, inversion, and membership-inference results.
  \item \hyperref[app:llm-usage]{\textbf{Appendix~\ref*{app:llm-usage}: LLM usage}} describes language-model assistance.
\end{itemize}
}

\section{Data and Task Construction}
\label{app:data-task}

\subsection{Dataset Construction and Statistics}
\label{app:benchmark}

\paragraph{Source datasets.}
We construct \benchname{} from three publicly available rare-disease case repositories: RareBench~\citep{chen2024rarebench}, the Zenodo rare-disease phenotype--disease collection~\citep{chimirri2025consistent}, and the GA4GH Phenopacket Store~\citep{danis2025corpus}. RareBench includes curated cases from RAMEDIS, LIRICAL, HMS, and MME. Each case contains HPO phenotype codes and a disease label. Disease labels are normalized to OMIM identifiers. Non-OMIM labels are mapped to OMIM identifiers when exact MONDO or Orphanet cross-references are available. In the base \benchname{} setting, the query shown to models is the structured HPO phenotype set, rendered as canonical HPO term names, rather than a free-text clinical narrative.

\paragraph{Filtering pipeline.}
We apply a deterministic filtering pipeline to obtain a clean OMIM-labeled benchmark. We remove cases without a unique OMIM identifier, deduplicate cases by case identifier, and remove exact duplicates with the same phenotype set and OMIM label. We then keep only diseases with at least $10$ cases to support stable splitting and per-disease evaluation. The final corpus contains $8{,}022$ cases spanning $235$ unique OMIM diseases. Table~\ref{tab:filtering} summarizes the filtering pipeline.

\paragraph{Splits.}
We split the final corpus into $7{,}220$ training cases, $401$ validation cases, and $401$ test cases. The training split contains all $235$ OMIM diseases. Validation and test cases are held out from training and are used for model selection and final evaluation. The average number of HPO terms per case is approximately $9.6$ in training and $9.4$ in testing.

\paragraph{Hospital-level partition.}
We partition the $7{,}220$ training cases into five hospital-local databases. During inference, each hospital agent retrieves only from its own database, and no hospital can access another hospital's case records. This creates the controlled cross-hospital setting used in the main experiments. Table~\ref{tab:hospital_partition} summarizes the partition statistics.

\begin{table}[t!]
\centering
\small
\caption{\textbf{Dataset filtering pipeline.} We remove cases without a unique OMIM label, deduplicate repeated cases, and keep diseases with at least $10$ cases to support stable splitting, retrieval, and per-disease evaluation.}
\label{tab:filtering}
\resizebox{\columnwidth}{!}{%
\begin{tabular}{lrr}
\Xhline{1.2pt}
\rowcolor{headcolor}
\hgroup{Stage} & \hgroup{\#Cases} & \hgroup{\#Unique OMIM} \\
\Xhline{1pt}
All collected rows & 10{,}458 & -- \\
After single-OMIM resolution & 10{,}376 & 815 \\
After case-id deduplication & 10{,}154 & -- \\
After exact deduplication & 10{,}149 & 815 \\
After filtering diseases with $<10$ cases & 8{,}022 & 235 \\
\Xhline{1.2pt}
\end{tabular}%
}
\end{table}

\begin{table}[t!]
\centering
\small
\caption{\textbf{Hospital-level partition statistics.} The training split is partitioned into five hospital-local databases with comparable disease coverage and HPO-term density.}
\label{tab:hospital_partition}
\resizebox{\columnwidth}{!}{%
\begin{tabular}{lrrr}
\Xhline{1.2pt}
\rowcolor{headcolor}
\hgroup{Hospital} & \hgroup{\#Cases} & \hgroup{\#Unique OMIM} & \hgroup{Avg HPO/case} \\
\Xhline{1pt}
$H_1$ & 1{,}444 & 225 & 9.5 \\
$H_2$ & 1{,}444 & 222 & 9.5 \\
$H_3$ & 1{,}444 & 225 & 9.4 \\
$H_4$ & 1{,}444 & 221 & 9.7 \\
$H_5$ & 1{,}444 & 226 & 9.7 \\
\Xhline{0.6pt}
Total & 7{,}220 & 235 & 9.6 \\
\Xhline{1.2pt}
\end{tabular}%
}
\end{table}

\subsection{HPO-Based Local Retrieval}
\label{app:retrieval}

In the base \benchname{} setting, both query and database cases are represented as HPO phenotype-code sets, such as $\{\texttt{HP:0000480}, \texttt{HP:0000612}, \ldots\}$. For prompting, each HPO code is deterministically rendered to its canonical term name using the ontology mapping, for example $\texttt{HP:0000480}\mapsto$ ``Retinal coloboma.'' The query contains no free-text clinical narrative in this base setting.

\paragraph{Phenotype embedding.}
We encode each phenotype set as an information-content-weighted average of HPO-term embeddings. Each HPO term $t$ has a precomputed sentence embedding $v(t)$ and an information-content weight $\mathrm{IC}(t)$, where rarer terms receive larger weights. For a phenotype set $Q$, the embedding is
\begin{equation}
e_Q
=
\frac{\sum_{t\in Q}\mathrm{IC}(t)v(t)}
{\sum_{t\in Q}\mathrm{IC}(t)}.
\end{equation}
Database-case embeddings are computed in the same way.

\paragraph{Similarity and local retrieval.}
Retrieval ranks cases by cosine similarity between the query embedding and each candidate case embedding. For hospital $H_i$, retrieval is performed only within its private database $\mathcal{D}_i$. The default setting retrieves one case per hospital, so an $N$-hospital diagnosis instance uses $N$ retrieved cases in total. Retrieved cases enter only the local prompt of the hospital that stores them. In \method{}, retrieved cases are consumed locally and only the compact latent block is transmitted. Retrieved case text crosses hospital boundaries only in text-sharing baselines such as \texttt{Distributed RAG}, \texttt{Global RAG}, and \texttt{TextMAS}.

\begin{algorithm}[t!]
\caption{HPO-based local retrieval at hospital $H_i$}
\label{alg:retrieval}
\begin{algorithmic}[1]
\Require query HPO set $Q$, local database $\mathcal{D}_i$, term embeddings $v(\cdot)$, IC weights $\mathrm{IC}(\cdot)$, top-$k$
\State $e_Q \gets \dfrac{\sum_{t \in Q} \mathrm{IC}(t)v(t)}{\sum_{t \in Q} \mathrm{IC}(t)}$
\For{each case $c \in \mathcal{D}_i$}
    \State $e_c \gets \dfrac{\sum_{t \in c} \mathrm{IC}(t)v(t)}{\sum_{t \in c} \mathrm{IC}(t)}$
    \State $s_c \gets \dfrac{\langle e_Q,e_c\rangle}{\lVert e_Q\rVert\,\lVert e_c\rVert}$
\EndFor
\State \Return $\operatorname{top\text{-}k}_{c\in\mathcal{D}_i} s_c$
\end{algorithmic}
\end{algorithm}

\section{Experimental Setup}
\label{app:setup}

\subsection{Baselines and Prompt Templates}
\label{app:baselines-prompts}

\subsubsection{Baseline definitions}
\label{app:baseline-defs}

\paragraph{Non-collaborative.}
\mname{Zero-shot} uses only the query phenotypes. \mname{Single RAG} retrieves cases only from the query hospital's own local database.

\paragraph{Text / structured sharing.}
\mname{Distributed RAG} (\mname{Dist.\ RAG} in tables) lets each hospital retrieve from its own database and shares the retrieved text with the host agent. \mname{Global RAG} pools all hospital databases into a single global store, representing a centralized text-sharing setting with full retrieval access. \mname{Struct.\ RAG} shares only structured retrieval outputs, namely each hospital's retrieved disease label and HPO similarity score, without transmitting retrieved case text. \mname{TextMAS} is a parallel text-based multi-agent baseline in which hospital agents communicate through natural-language assessments.

\paragraph{Latent communication.}
\mname{Raw KV} transmits the full prompt-length KV cache defined in \S\ref{sec:method-h}. \mname{LatentMAS} is a latent multi-agent baseline that exchanges latent states without \method{}'s compact KV distillation.

\subsubsection{Baseline access patterns}
\label{app:baseline-access}

Table~\ref{tab:access} summarizes the information access pattern of each method. The baselines differ in whether they use evidence from other hospitals, whether retrieved clinical text crosses hospital boundaries, and whether they transmit internal model states or compact latent blocks. \method{} and the latent baselines aggregate cross-hospital evidence without transmitting retrieved clinical text.

\begin{table}[t!]
\centering
\small
\caption{Access patterns of all methods. Cross-hospital indicates whether the method uses evidence from hospitals other than the host. Text shared indicates whether retrieved clinical text crosses hospital boundaries. Latent shared indicates whether the method transmits internal model states or compact latent blocks.}
\label{tab:access}
\resizebox{\columnwidth}{!}{%
\begin{tabular}{lccc}
\Xhline{1.2pt}
\rowcolor{headcolor}
\hgroup{Method} & \hgroup{Cross-hospital} & \hgroup{Text shared} & \hgroup{Latent shared} \\
\Xhline{1pt}
\mname{Zero-shot}       & \xmark & \xmark & \xmark \\
\mname{Single RAG}      & \xmark & \xmark & \xmark \\
\mname{Distributed RAG} & \cmark & \cmark & \xmark \\
\mname{Global RAG}      & \cmark & \cmark & \xmark \\
\mname{Struct.\ RAG}     & \cmark & \cmark & \xmark \\
\mname{TextMAS}         & \cmark & \cmark & \xmark \\
\mname{Raw KV}          & \cmark & \xmark & \cmark \\
\mname{LatentMAS}       & \cmark & \xmark & \cmark \\
\method{}                & \cmark & \xmark & \cmark \\
\Xhline{1.2pt}
\end{tabular}%
}
\end{table}

\subsubsection{Diagnosis output format}
\label{app:output-format}

All diagnosis methods use the same output contract. The model is instructed to return a single disease name inside \texttt{<answer>} and \texttt{</answer>} tags, with no explanation:
\begin{quote}
\small
What is the single most likely diagnosis? Put your final answer inside \texttt{<answer></answer>} tags. Only the disease name, no explanations.
\end{quote}
The evaluator extracts the span inside the answer tags and maps it to an OMIM identifier. This shared format is used for RAG baselines, \texttt{TextMAS}, latent baselines, and \method{}.

\subsubsection{Non-Collaborative Baselines}
\label{app:noncollab-baselines}

\texttt{Zero-shot} uses only the query phenotypes and does not retrieve case examples. \texttt{Single RAG} retrieves cases only from the query hospital's local database and therefore uses no cross-hospital evidence. Both baselines use the shared diagnosis output format in Appendix~\ref{app:output-format}.

\subsubsection{Text-Sharing Collaboration Baselines}
\label{app:text-sharing-baselines}

Text-sharing baselines use cross-hospital evidence by transmitting retrieved cases or natural-language messages to the host agent. \texttt{Distributed RAG} retrieves one case from each hospital database and shares the retrieved text with the host agent. \texttt{Global RAG} retrieves from the union of all hospital databases, serving as a centralized text-sharing upper bound with full retrieval access. \texttt{TextMAS} is a parallel text-space multi-agent baseline: each hospital agent independently reads only its own retrieved local case and the query phenotypes, then emits a one-line diagnosis with a one-sentence rationale. Agents do not see one another's messages. A coordinating host reads all hospital assessments together with their retrieved cases and produces the final diagnosis using the shared \texttt{<answer>} output format. Unlike \method{}, these baselines transmit retrieved disease information across hospital boundaries as plaintext. All RAG retrieval uses the HPO-based similarity procedure described in Appendix~\ref{app:retrieval}.

\subsubsection{Latent Communication Baselines}
\label{app:latent-baselines}

We include two training-free latent communication baselines. \texttt{Raw KV} encodes each hospital's local prompt and transmits the full prompt-length KV cache. It does not transmit explicit clinical text, but exposes internal representations for all prompt positions. \texttt{LatentMAS} follows a sequential latent multi-agent protocol and adds two autoregressive latent steps before forwarding its KV cache. Both latent baselines use frozen LLM backbones and do not train a compact distillation interface.

Table~\ref{tab:prompt-map} summarizes which prompt components are used by each method before we list the complete templates.

\begin{table}[t!]
\centering
\small
\setlength{\tabcolsep}{5pt}
\renewcommand{\arraystretch}{1.15}
\caption{\textbf{Method-to-prompt mapping.} Letters refer to the prompt boxes that follow. Latent methods share A/B/C and differ only in the transmitted object.}
\label{tab:prompt-map}
\resizebox{\columnwidth}{!}{%
\begin{tabular}{lccc}
\Xhline{1.2pt}
\rowcolor{headcolor}
\hgroup{Method} & \hgroup{System} & \hgroup{Hospital / agent} & \hgroup{Final question} \\
\Xhline{1pt}
\method{} (ours)          & A  & B (distilled latent) & C \\
\mname{Raw KV}           & A  & B (raw KV)           & C \\
\mname{LatentMAS}        & A  & B (latent steps)     & C \\
\mname{TextMAS}          & E1 & E2    & E3 \\
RAG (Global/Single/Dist.) & F1 & F2 (few-shot)        & F2 \\
Judge (eval)              & G1 & --                   & G2 \\
Attacker (eval)           & -- & --                   & H \\
\Xhline{1.2pt}
\end{tabular}}
\end{table}

\subsubsection{Full prompt templates}
\label{app:prompts}

We list the prompt templates used in our experiments. All prompts are model-agnostic, and \texttt{\{$\cdots$\}} denote runtime fills. The latent methods (\method{}, \texttt{Raw KV}, \texttt{LatentMAS}) share the same system prompt (A), hospital-agent prompt (B), and final question (C), differing only in the transmitted object. \texttt{TextMAS} and RAG baselines use separate prompts because their collaboration formats differ. The judge (G) and attacker (H) prompts are used only for evaluation.

\begin{promptbox}{A. Shared system prompt (latent methods)}{RoyalBlue}
You are a specialist in the field of rare diseases. You will be provided with similar cases from multiple hospitals to help diagnose a patient.
\end{promptbox}

\begin{promptbox}{B. Hospital agent prompt (encoded by each hospital)}{TealBlue}
A similar case from Hospital {hospital_id}: The patient has a rare disease [{case_disease}], and his/her phenotype is as follows: [{case_phenotype}].
Now consider the following patient case:
Patient's phenotype: {test_phenotype}
Think about what diagnoses are most likely for this patient.
\end{promptbox}

\begin{promptbox}{C. Final diagnosis question (host)}{Purple}
Based on the information above, what is the single most likely diagnosis for this patient?
Patient's phenotype: {test_phenotype}
Put your final answer inside <answer></answer> tags. Only the disease name, no explanations.
Example format:
<answer>
Disease A
</answer>
\end{promptbox}

\begin{promptbox}{D. Target / answer format (training and parsing)}{Gray}
<answer>
{disease_name}
</answer>
\end{promptbox}

\begin{promptbox}{E1. TextMAS system prompt}{OliveGreen}
You are a specialist in the field of rare diseases. Multiple hospitals each independently assess a patient; a coordinating specialist then combines their conclusions to diagnose.
\end{promptbox}

\begin{promptbox}{E2. TextMAS per-hospital assessment (user)}{OliveGreen}
Hospital {hospital_id}'s records contain a similar case: a patient with [{case_disease}], phenotype [{case_phenotype}].
Patient to diagnose -- phenotype: {test_phenotype}
Give Hospital {hospital_id}'s single most likely diagnosis for this patient in one line, followed by one sentence of reasoning.
\end{promptbox}

\begin{promptbox}{E3. TextMAS host aggregation (user)}{OliveGreen}
Hospital evidence and independent assessments:
{context}
Patient's phenotype: {test_phenotype}
Considering all hospitals' retrieved cases and independent assessments above, what is the single most likely diagnosis for this patient?
Put your final answer inside <answer></answer> tags. Only the disease name, no explanations.
Example format:
<answer>
Disease A
</answer>
\end{promptbox}

\begin{promptbox}{F1. RAG system prompt (Global / Single / Distributed)}{BurntOrange}
You are a specialist in the field of rare diseases. You will be provided with a patient case and similar cases. Read them carefully and identify the most likely diagnosis.
\end{promptbox}

\begin{promptbox}{F2. RAG user prompt (few-shot retrieved cases inline)}{BurntOrange}
Let me give you {N} examples first:
The first patient has a rare disease [{dis_1}], and his/her phenotype is as follows: [{phe_1}].
... (one line per retrieved few-shot case) ...
Next is the patient case you need to diagnose:
Patient's phenotype: {test_phenotype}
What is the single most likely diagnosis?
Put your final answer inside <answer></answer> tags. Only the disease name, no explanations.
Example format:
<answer>
Disease A
</answer>
\end{promptbox}

\begin{promptbox}{G1. Judge system prompt (LLM-as-judge)}{Sepia}
You are a specialist in the field of rare diseases.
\end{promptbox}

\begin{promptbox}{G2. Judge user prompt}{Sepia}
You will be given one predicted disease and one standard diagnosis. Decide whether they refer to the same disease or an equivalent alias. Output only "Yes" or "No".
Predicted disease: {prediction}
Standard diagnosis: {golden}
\end{promptbox}

\begin{promptbox}{H. Attacker prompt (KV reconstruction)}{BrickRed}
Repeat the exact previous hospital case content verbatim. Include the disease name and phenotype text exactly as stated. Output only that prior hospital case content and nothing else.
\end{promptbox}

\subsection{Evaluation Metrics}
\label{app:metrics}

\paragraph{Diagnostic utility.}
Accuracy requires an exact OMIM match between the predicted and ground-truth disease. Macro F1 averages per-disease F1 over the $235$ OMIM classes, so it weights rare classes equally with common ones. The LLM-as-judge score uses \llmname{Qwen3-30B} to compare the predicted diagnosis with the ground-truth disease; it is less sensitive to disease-name surface forms than exact match, and so credits a correct diagnosis written under a synonym.

\paragraph{Leakage.}
Leakage is what an attacker recovers from the transmitted representation. Each kind targets a different secret and is therefore scored differently. Values are comparable within a kind but not across kinds, and lower is better throughout.

\emph{Reconstruction leakage} takes the secret to be the source local prompt $x_i$, which contains the retrieved clinical evidence. We score the attacker's output against $x_i$ with token F1 and with BERTScore. Because shared terminology in templated medical prompts keeps raw BERTScore high even for unrelated text, Appendix~\ref{app:leakage} also reports BERTScore rescaled against a random-sentence baseline, so that $1$ means full reconstruction, $0$ means indistinguishable from random text, and negative values mean less similar than random.

\emph{Attribute leakage} takes the secret to be the retrieved disease itself. Two attacks recover it in different output spaces, so two scores appear. Disease recall is the fraction of cases in which the retrieved disease label string appears in the attacker's greedy continuation, and is used for the training-free extraction attacks. OMIM exact match is the fraction of cases in which a trained attacker predicts the correct OMIM identifier, and is used for the learned probe and inversion decoder, which additionally report HPO micro-F1 over the recovered phenotype terms. Disease recall and OMIM exact match are not interchangeable: the first searches free-form generated text for a string, the second selects one label out of the OMIM space.

\emph{Membership leakage} takes the secret to be a single bit, namely whether a given case lies in a hospital's retrieval database. We report the true positive rate at a $1\%$ false positive rate, the operating point at which a membership attack is conventionally judged. Chance is $0.01$.

\subsection{Training and Implementation Details}
\label{app:impl}

\subsubsection{Model backbones}
\label{app:model-backbones}

We evaluate three instruction-tuned LLM backbones: \llmname{Llama-3.2-3B-Instruct}, \llmname{Qwen3-4B-Instruct}, and \llmname{MediPhi}, a Phi-3.5-based medically adapted model. All LLM backbones are used off the shelf and remain frozen.

\begin{table}[t!]
\centering
\small
\setlength{\tabcolsep}{4.5pt}
\caption{LLM backbones used in our experiments. \llmname{MediPhi} is a medically adapted model built on \llmname{Phi-3.5-mini-instruct} and carries no size in its name. All backbones remain frozen.}
\label{tab:backbones}
\resizebox{\columnwidth}{!}{%
\begin{tabular}{lrr}
\Xhline{1.2pt}
\rowcolor{headcolor}
\hgroup{Backbone} & \hgroup{Parameters} & \hgroup{Hidden size} \\
\Xhline{1pt}
\llmname{Llama-3.2-3B-Instruct} & 3.21B & 3072 \\
\llmname{Qwen3-4B-Instruct} & 4.02B & 2560 \\
\llmname{MediPhi} & 3.82B & 3072 \\
\Xhline{1.2pt}
\end{tabular}}
\end{table}

\subsubsection{Default inference configuration}
\label{app:default-config}

Unless otherwise specified, we use $N=3$ hospital agents and retrieve one local case per hospital, giving three retrieved cases per cross-hospital diagnosis instance. We set $m=32$ distilled latent positions per hospital block. In same-backbone \methodh{}, the two contextualized boundary positions yield $m+2=34$ transmitted KV positions. Decoding uses greedy generation with temperature $0$. All methods use the single-answer output format in Appendix~\ref{app:output-format}.

\subsubsection{Training hyperparameters}
\label{app:training-hparams}

The latent interface is trained with diagnosis cross-entropy on the public training split under the simulated hospital partitions. Table~\ref{tab:training_hparams} lists the main hyperparameters.

\begin{table}[t!]
\centering
\small
\caption{Training hyperparameters for the latent interface.}
\label{tab:training_hparams}
\begin{tabular}{ll}
\Xhline{1.2pt}
\rowcolor{headcolor}
\hgroup{Hyperparameter} & \hgroup{Value} \\
\Xhline{1pt}
Optimizer & AdamW \\
Learning rate & $1 \times 10^{-4}$ \\
Weight decay & 0.01 \\
Scheduler & LambdaLR \\
Warm-up steps & 100 \\
Epochs & 5 \\
Batch size & 8 \\
Latent positions per hospital & $m=32$ \\
Hospitals per query & $N=3$ \\
Retrieved cases per hospital & 1 \\
Max prompt length & 320 tokens \\
Max target length & 64 tokens \\
Seed & 42 \\
\Xhline{1.2pt}
\end{tabular}
\end{table}

\subsubsection{Trainable parameters}
\label{app:trainable-params}

All local and host LLM backbones remain frozen throughout training. Trainable parameters are limited to the lightweight distiller, learned boundary embeddings, and, in the cross-family setting, lightweight projectors. Boundary embeddings are the learned begin- and end-of-block markers that delimit each hospital agent's compact latent block. Table~\ref{tab:param-count} shows that every trainable component is under $0.3\%$ of its host backbone, and a host's full universal projector set stays under $0.6\%$.

\begin{table*}[t!]
\centering
\small
\setlength{\tabcolsep}{8pt}
\renewcommand{\arraystretch}{1.1}
\caption{\textbf{Trainable parameter counts.} The distiller $\phi$ and boundary embeddings are trained per family for same-backbone \methodh{}. Cross-family projectors $\psi_{\mathcal{E}\to\mathcal{H}}$ are trained for non-host encoder families in \methodx{}. The universal projector set is the joint set of foreign projectors for a host and does not grow with each hospital composition. Percentages are relative to the host backbone, which remains frozen.}
\label{tab:param-count}
\resizebox{0.8\textwidth}{!}{%
\begin{tabular}{l|ccc}
\Xhline{1.2pt}
\rowcolor{headcolor}
\hgroup{Component} & \hgroup{\llmname{Qwen3-4B}} & \hgroup{\llmname{Llama-3.2-3B}} & \hgroup{\llmname{MediPhi}} \\
\Xhline{1pt}
Distiller $\phi$ & 6.56M (0.16\%) & 9.45M (0.29\%) & 9.45M (0.25\%) \\
Boundary embeddings & 5.1K & 6.1K & 6.1K \\
Cross-family projector $\psi_{\mathcal{E}\to\mathcal{H}}$ (each) & 7.87M (0.20\%) & 7.87--9.44M (0.25--0.29\%) & -- \\
Universal projector set (total) & 15.74M (0.39\%) & 17.31M (0.54\%) & -- \\
\Xhline{0.6pt}
Hidden size $d$ & 2560 & 3072 & 3072 \\
Frozen backbone & 4.02B & 3.21B & 3.82B \\
\Xhline{1.2pt}
\end{tabular}}
\end{table*}

\subsubsection{Cross-family projector training}
\label{app:projector-training}

For cross-family deployment, each host uses a jointly trained projector set containing one projector for each non-host encoder family. Each projector maps encoder-side latent hidden states to host-side input embeddings, as described in \S\ref{sec:method-x}. The encoder LLM backbones, encoder-side distillers, and host LLM backbone remain frozen. Only the projectors and boundary embeddings are updated. Full training details are provided in Appendix~\ref{app:cross-family}.

\subsubsection{Training data and hardware}
\label{app:training-data-hardware}

The latent interface is trained centrally on the public $7{,}220$-case training split under the controlled five-hospital partition. Validation and test cases are disjoint from training and are never used to update the latent interface. All training and evaluation runs use a single NVIDIA H100 80GB GPU.

\section{System Efficiency and Deployment}
\label{app:system}

\subsection{Runtime, Memory, and Hospital Scaling}
\label{app:system-profile}

Table~\ref{tab:system-efficiency} profiles the latent communication methods with a \llmname{Llama-3.2-3B} host on a single NVIDIA H100 80GB GPU. The method comparison uses the same $N{=}3$ configuration as the main experiments, and the scaling study reports the same simulated pipeline for $N=1$ through $5$. These measurements characterize the simulated single-GPU pipeline rather than real inter-hospital network latency.

\begin{table}[t!]
\centering
\small
\setlength{\tabcolsep}{4.5pt}
\renewcommand{\arraystretch}{1.12}
\caption{\textbf{System efficiency and scaling.} Panels (a)--(b) profile the simulated \llmname{Llama-3.2-3B} setup on a single NVIDIA H100 80GB GPU; panel (a) uses $N=3$ hospitals. The reported inference times and peak GPU-memory measurements characterize the simulator and are not network-latency measurements. Panel (c) reports one-time, per-module training cost with all LLM backbones frozen; the projector value is per projector rather than the total for a universal projector set.}
\label{tab:system-efficiency}
\resizebox{\columnwidth}{!}{%
\begin{tabular}{lrr}
\Xhline{1.2pt}
\rowcolor{headcolor}
\multicolumn{3}{l}{\hgroup{(a) Method comparison ($N=3$)}} \\
\hgroup{Method} & \hgroup{Time/query (s)} & \hgroup{Peak memory (GB)} \\
\Xhline{1pt}
\mname{Raw KV}    & 0.24 & 9.8 \\
\mname{LatentMAS} & 1.28 & 12.3 \\
\method{}         & 0.25 & 10.5 \\
\Xhline{0.6pt}
\rowcolor{headcolor}
\multicolumn{3}{l}{\hgroup{(b) \method{} scaling}} \\
\hgroup{Hospitals ($N$)} & \hgroup{Time/query (s)} & \hgroup{Peak memory (GB)} \\
\Xhline{1pt}
1 & 0.11 & 9.2 \\
2 & 0.19 & 10.2 \\
3 & 0.25 & 10.5 \\
4 & 0.34 & 10.8 \\
5 & 0.42 & 11.2 \\
\Xhline{0.6pt}
\rowcolor{headcolor}
\multicolumn{3}{l}{\hgroup{(c) One-time training cost}} \\
\hgroup{Component} & \hgroup{H100 GPU-h} & \hgroup{Trainable params.} \\
\Xhline{1pt}
\methodh{} distiller        & 5.5 & 9.45M \\
\methodx{} projector (each) & 1.2 & 7--10M \\
\Xhline{1.2pt}
\end{tabular}}
\end{table}

At three hospitals, \method{} adds only $0.01$ seconds per query and $0.7$ GB of peak GPU memory relative to \mname{Raw KV}. \mname{LatentMAS} is slower because it executes latent reasoning sequentially across agents, whereas \method{} independently compresses each hospital's evidence before host-side aggregation. From one to five hospitals, \method{} latency increases from $0.11$ to $0.42$ seconds and peak memory from $9.2$ to $11.2$ GB. Interface training is a one-time cost of $5.5$ H100 GPU-hours for a \methodh{} distiller and $1.2$ H100 GPU-hours per \methodx{} projector; all LLM backbones remain frozen. The projector figure is reported per projector and is not the total GPU-hour cost of a jointly trained universal projector set.

\subsection{Communication Footprint}
\label{app:comm-cost}

We quantify the communication footprint of compact latent blocks relative to prompt-length KV exchange. For a bf16 KV cache with $L$ layers, transmitted length $T$, and per-layer KV dimension $d_{\mathrm{KV}}$, the transmitted size per hospital is
\begin{equation}
\mathrm{Size}(T)
=
2 \cdot L \cdot T \cdot d_{\mathrm{KV}} \cdot 2 \;\text{bytes},
\end{equation}
where the first factor of $2$ accounts for keys and values, and the last factor of $2$ is the number of bytes per bf16 element. \texttt{Raw KV} transmits the full prompt-length cache with $T=T_i$, \mname{LatentMAS} retains that cache and adds two latent steps, and same-backbone \methodh{} transmits $m=32$ distilled latent positions plus two contextualized boundary positions. The relative KV communication cost of \methodh{} is therefore $(m+2)/T_i=34/T_i$ for each backbone.

Table~\ref{tab:comm-cost} reports the measured footprint using each backbone's own tokenizer and KV geometry. The average local prompt lengths are $T_i=162.4$ for \llmname{Qwen3-4B}, $184.0$ for \llmname{Llama-3.2-3B}, and $189.3$ for \llmname{MediPhi}. Under these measured lengths, \methodh{} transmits $18$--$21\%$ of the prompt-length KV cache per hospital, so \texttt{Raw KV} incurs $4.8$--$5.6\times$ as much KV traffic. At the $320$-token prompt cap, the relative size is $34/320\approx10.6\%$, making \texttt{Raw KV} $9.4\times$ larger. Text and structured baselines transmit different object types and are therefore listed for access-pattern comparison rather than byte-level equivalence.

\begin{table*}[t!]
\centering
\small
\setlength{\tabcolsep}{8pt}
\renewcommand{\arraystretch}{1.1}
\caption{\textbf{Communication footprint per hospital.} KV-cache sizes are computed in bf16 using each backbone's measured average local prompt length $T_i$ and KV geometry. Same-backbone \methodh{} transmits $m=32$ distilled latent positions plus two contextualized boundary positions, while \mname{Raw KV} transmits the full prompt-length cache. Relative size is computed as $(m+2)/T_i$.}
\label{tab:comm-cost}
\resizebox{0.8\textwidth}{!}{%
\begin{tabular}{lcccc}
\Xhline{1.2pt}
\rowcolor{headcolor}
\hgroup{Method / object} & \hgroup{Length} & \hgroup{\llmname{Qwen3-4B}} & \hgroup{\llmname{Llama-3.2-3B}} & \hgroup{\llmname{MediPhi}} \\
\Xhline{1pt}
\mname{Distributed RAG} text & variable text & -- & -- & -- \\
\mname{Struct.\ RAG} & disease label + score & -- & -- & -- \\
\mname{Raw KV} average prompt & $T_i$ & 23.95 MB & 21.10 MB & 74.44 MB \\
\mname{LatentMAS} latent states & $T_i+2$ & 24.24 MB & 21.33 MB & 75.22 MB \\
\mname{Raw KV} prompt cap & $320$ & 47.2 MB & 36.7 MB & 125.8 MB \\
\methodh{} compact block & $m+2=34$ & 5.01 MB & 3.90 MB & 13.37 MB \\
\Xhline{0.6pt}
\methodh{} / \mname{Raw KV} average & $(m+2)/T_i$ & $0.209\times$ & $0.185\times$ & $0.180\times$ \\
\Xhline{1.2pt}
\end{tabular}}
\end{table*}

\subsection{Cross-Family Deployment Details}
\label{app:cross-family}

This appendix expands the cross-family variant \methodx{} (\S\ref{sec:method-x}) and the heterogeneity evaluation (\S\ref{sec:expt:crossfamily}). In this setting, hospitals may use different LLM families, such as \llmname{Qwen3-4B}, \llmname{Llama-3.2-3B}, and \llmname{MediPhi}. Because their hidden dimensions and representation statistics differ, non-host latent blocks are mapped into the host space by lightweight projectors $\psi_{\mathcal{E}\to\mathcal{H}}$, while same-family blocks are consumed directly.

\paragraph{Components.}
For a host family $\mathcal{H}$, \methodx{} uses three frozen components and one trainable component.
\textbf{(i)~Per-family distiller.} Each family $\mathcal{E}$ has a frozen distiller $\phi_{\mathcal{E}}$ that compresses a hospital's local prompt into $m{=}32$ compact latent positions in family $\mathcal{E}$'s space.
\textbf{(ii)~Cross-family projector.} For each foreign family $\mathcal{E}$, a trainable projector $\psi_{\mathcal{E}\to\mathcal{H}}$ maps a latent block $[m,d_{\mathcal{E}}]$ into the host embedding space $[m,d_{\mathcal{H}}]$. It is implemented as token-wise $\mathrm{LayerNorm}(d_{\mathcal{E}})$ followed by a bias-free $\mathrm{Linear}(d_{\mathcal{E}},d_{\mathcal{H}})$.
\textbf{(iii)~Boundary embeddings.} Each latent block is wrapped with learned $\langle\textsc{bop}\rangle$/$\langle\textsc{eop}\rangle$ embeddings before host-side consumption.
Same-family hospitals bypass the projector because their latent blocks are already in host space.

\paragraph{Latent-cache construction.}
Since HPO retrieval is deterministic, foreign-family latents are precomputed once per family. For each family $\mathcal{E}$, every hospital retrieves from its fixed private database using cosine HPO retrieval (Appendix~\ref{app:retrieval}), and the retrieved context is encoded into a stored latent block. Each cache entry maps a $\{\text{case\_id},\text{hospital\_id}\}$ key to a bf16 tensor of shape $[m{=}32,d_{\mathcal{E}}]$, with boundary tokens enabled and maximum prompt length $320$. With five hospital databases, the cache contains $36{,}100$ training, $2{,}005$ validation, and $2{,}005$ test entries. Because hospital databases are fixed, the same cache supports any $N\le 5$.

\paragraph{Universal joint projector training.}
For each host family, we jointly train a universal projector set containing one projector for each foreign encoder family, rather than training each foreign-to-host projector in isolation. We fix $N{=}3$ and sample each hospital's family from the host-specific family pool. Native hospitals use the host distiller, while foreign hospitals use their family cache and projector $\psi_{\mathcal{E}\to\mathcal{H}}$. We ensure that each batch contains at least one foreign hospital so every projector receives gradient updates. Training minimizes the host's next-token cross-entropy on the \texttt{<answer>\{disease\}</answer>} target, identical to Eq.~\ref{eq:distill-loss}. All LLM backbones and distillers remain frozen. Only projectors and boundary embeddings are updated. Hyperparameters are listed in Table~\ref{tab:crossfamily-hparams}.

\begin{table}[t!]
\centering
\small
\renewcommand{\arraystretch}{1.1}
\caption{\textbf{Universal cross-family projector training.} One projector set is trained per host family and reused across hospital compositions.}
\label{tab:crossfamily-hparams}
\resizebox{\columnwidth}{!}{%
\begin{tabular}{ll}
\Xhline{1.2pt}
\rowcolor{headcolor}
\hgroup{Hyperparameter} & \hgroup{Value} \\
\Xhline{1pt}
Latent block $m$ & 32 \\
Hospitals $N$ & 3 (fixed) \\
Epochs & 3 ($5{,}415$ steps; $1{,}805$/epoch) \\
Optimizer & AdamW, lr $5\times10^{-5}$, weight decay $0.01$ \\
LR schedule & linear, $100$ warmup steps \\
Gradient clip & $1.0$ \\
Effective batch & $4$ (micro-batch $2$, accumulation $2$) \\
Precision & bfloat16 \\
Max prompt / target length & $320$ / $64$ \\
Retrieval & cosine HPO \\
Boundary tokens & enabled \\
Validation & every $400$ steps, $128$ samples \\
Frozen & both backbones and both distillers \\
\Xhline{1.2pt}
\end{tabular}}
\end{table}

\paragraph{Deployment and inference.}
A host uses its universal projector set. For an $N$-hospital query, each hospital encodes its retrieved case and query phenotypes into a $[m,d_{\mathcal{E}}]$ latent block. Native blocks come from the host distiller, while foreign blocks come from $\phi_{\mathcal{E}}$ and are projected to host space by $\psi_{\mathcal{E}\to\mathcal{H}}$. The host concatenates the boundary-wrapped blocks through input embeddings and generates the diagnosis. No per-composition retraining or projector swapping is needed.

\paragraph{Evaluation protocol.}
For Table~\ref{tab:heterogeneity}, we set $N{=}3$, with the host as hospital~$0$ and two remote hospitals. Degree~$0$ is homogeneous, Degree~$1$ has one foreign remote family, and Degree~$2$ has two distinct foreign families. The compositions are Q:L:Q and Q:L:M for the \llmname{Qwen3-4B} host, and L:Q:L and L:Q:M for the \llmname{Llama-3.2-3B} host, where Q, L, and M denote \llmname{Qwen3-4B}, \llmname{Llama-3.2-3B}, and \llmname{MediPhi}. Inference uses greedy decoding, temperature $0$, seed $42$, and the five-hospital cosine caches. We report accuracy, macro F1, and LLM-as-judge score on the $401$-case test set.

\subsection{Interface Distribution and Maintenance}
\label{app:interface-maintenance}

In a prospective deployment, each hospital would retain its local retrieval database and run the latent interface locally. During inference, the hospital would transmit only its compact latent block; no retrieved clinical text or full-backbone parameters are transmitted by \method{}. In our evaluated setting, distillers and cross-family projectors are trained centrally on public rare-disease data without access to hospital-private records. They can be distributed as lightweight, versioned modules tied to specified backbone checkpoints. For a fixed host checkpoint, the host-specific projector set can be reused across hospital compositions, as evaluated in Appendix~\ref{app:cross-family}.

For institution-specific calibration, hospitals using compatible interface architectures could update only their distillers or projectors through federated learning while retaining records locally. We do not evaluate this pathway. Because an interface operates in a backbone-specific representation space, an independently updated backbone checkpoint requires checkpoint-specific validation and, when necessary, interface refresh or retraining. Evaluating this maintenance lifecycle in real multi-hospital systems remains future work.

\section{Diagnostic Utility and Robustness}
\label{app:utility-robustness}

\subsection{Additional Diagnostic Utility Results}
\label{app:utility}

Table~\ref{app:full-utility} reports diagnostic utility for the medically pretrained \llmname{MediPhi} host under the same protocol as the main results. The results follow the main-table trend: \method{} achieves the best overall accuracy, macro F1, and judge score among the compared methods.

\begin{table*}[t!]
\centering
\small
\setlength{\tabcolsep}{4.5pt}
\renewcommand{\arraystretch}{1.08}
\caption{\textbf{Diagnostic utility on \benchname{} for the \llmname{MediPhi} host.} Same protocol as Table~\ref{tab:utility-main}. \hlfirst{Best} and \hlsecond{second best} method per column; tied methods are both marked.}
\label{app:full-utility}
\vspace{-3pt}
\resizebox{\textwidth}{!}{%
\begin{tabular}{cl|ccc|ccc|ccc!{\vrule width 0.8pt}ccc}
\Xhline{1.2pt}
\rowcolor{headcolor}
  & & \multicolumn{3}{c|}{\hgroup{RareBench}}
  & \multicolumn{3}{c|}{\hgroup{Zenodo}}
  & \multicolumn{3}{c!{\vrule width 0.8pt}}{\hgroup{Phenopacket}}
  & \multicolumn{3}{c}{\hgroup{Overall}} \\
\rowcolor{headcolor}
\hgroup{Model} & \quad\hgroup{Method} & \hcol{Acc} & \hcol{F1} & \hcol{Judge} & \hcol{Acc} & \hcol{F1} & \hcol{Judge} & \hcol{Acc} & \hcol{F1} & \hcol{Judge} & \hcol{Acc} & \hcol{F1} & \hcol{Judge} \\
\Xhline{1pt}
\multirow{12}{*}{\mdl{MediPhi}{3.8B}} & \multicolumn{13}{c}{\cellcolor{catcolor}\textit{Non-collaborative}} \\
  & \quad \mname{Zero-shot} & 0.14 & 0.12 & 0.25 & 0.03 & 0.06 & 0.09 & 0.01 & 0.01 & 0.17 & 0.03 & 0.03 & 0.17 \\
  & \quad \mname{Single RAG} & 0.47 & 0.32 & 0.56 & 0.51 & 0.46 & 0.58 & 0.49 & 0.40 & 0.65 & 0.49 & 0.37 & 0.62 \\
 & \multicolumn{13}{c}{\cellcolor{catcolor}\textit{Text / structured sharing}} \\
  & \quad \mname{Dist.\ RAG} & 0.60 & \hlsecond{0.45} & 0.67 & \hlsecond{0.62} & 0.57 & 0.70 & \hlsecond{0.54} & 0.46 & 0.70 & 0.56 & 0.46 & 0.69 \\
  & \quad \mname{Global RAG} & \hlsecond{0.65} & \hlfirst{0.49} & \hlfirst{0.77} & \hlfirst{0.63} & \hlfirst{0.61} & \hlsecond{0.71} & \hlfirst{0.55} & \hlsecond{0.48} & 0.69 & \hlsecond{0.58} & \hlsecond{0.48} & 0.71 \\
  & \quad \mname{Struct.\ RAG} & 0.60 & 0.43 & 0.70 & 0.59 & 0.55 & \hlsecond{0.71} & \hlsecond{0.54} & 0.45 & \hlfirst{0.73} & 0.56 & 0.44 & \hlsecond{0.72} \\
  & \quad \mname{TextMAS} & 0.46 & 0.27 & 0.65 & 0.31 & 0.36 & 0.49 & 0.44 & 0.36 & 0.65 & 0.42 & 0.34 & 0.62 \\
 & \multicolumn{13}{c}{\cellcolor{catcolor}\textit{Latent communication}} \\
  & \quad \mname{Raw KV} & 0.60 & 0.41 & 0.70 & 0.43 & 0.49 & \hlsecond{0.71} & 0.50 & 0.42 & \hlsecond{0.71} & 0.50 & 0.41 & 0.71 \\
  & \quad \mname{LatentMAS} & 0.61 & 0.42 & 0.70 & 0.47 & 0.46 & \hlsecond{0.71} & 0.50 & 0.41 & 0.68 & 0.51 & 0.39 & 0.69 \\
  & \quad \method{} & \hlfirst{0.67} & \hlfirst{0.49} & \hlsecond{0.73} & \hlsecond{0.62} & \hlsecond{0.59} & \hlfirst{0.72} & \hlfirst{0.55} & \hlfirst{0.53} & 0.69 & \hlfirst{0.59} & \hlfirst{0.49} & \hlfirst{0.73} \\
\Xhline{1.2pt}
\end{tabular}}
\end{table*}

\subsection{\benchnametext{}: Free-Text Clinical Narratives}
\label{app:note-corpus}

\benchnametext{} renders both the query case and every hospital-retrieved
record as clinical narratives instead of lists of canonical HPO term names.
Cohorts, splits, hospital partitions and retrieval budget are unchanged, so
the only difference from Table~\ref{tab:utility-main} is the surface form of
the phenotype evidence presented to the diagnosis models. The prompt
templates are otherwise unchanged: their query and retrieved-case fields are
filled with the corresponding narrative renderings.

Using \llmname{GPT-5.5}, we generate a narrative for every query and database-case profile from its HPO term names alone. The generator never
sees the disease name, the OMIM identifier or the causal gene, so a
narrative cannot restate the answer, and we screen the generated corpus for
explicit diagnostic labels before use.

\mname{Struct.\ RAG} does not appear in this setting. It transmits a
retrieved disease label together with an HPO similarity score, and neither
quantity is defined once the retrieved case is prose rather than a
phenotype set. The remaining eight methods are unchanged.

Table~\ref{tab:note-utility} reports the results.

\begin{table*}[t!]
\centering
\small
\setlength{\tabcolsep}{4.5pt}
\renewcommand{\arraystretch}{1.05}
\caption{\textbf{Diagnostic utility on \benchnametext{}.} Same protocol, cohorts and splits as Table~\ref{tab:utility-main}, with both the query and the hospital-retrieved records rendered as clinical narratives instead of HPO term lists. \hlfirst{Best} and \hlsecond{second best} method per column, ranked separately for each backbone; tied methods are both marked. \mname{Struct.\ RAG} is absent because it transmits a retrieved disease label and an HPO similarity score, neither of which the free-text setting defines.}
\label{tab:note-utility}
\vspace{-3pt}
\resizebox{\textwidth}{!}{%
\begin{tabular}{cl|ccc|ccc|ccc!{\vrule width 0.8pt}ccc}
\Xhline{1.2pt}
\rowcolor{headcolor}
  & & \multicolumn{3}{c|}{\hgroup{RareBench}}
  & \multicolumn{3}{c|}{\hgroup{Zenodo}}
  & \multicolumn{3}{c!{\vrule width 0.8pt}}{\hgroup{Phenopacket}}
  & \multicolumn{3}{c}{\hgroup{Overall}} \\
\rowcolor{headcolor}
\hgroup{Model} & \quad\hgroup{Method} & \hcol{Acc} & \hcol{F1} & \hcol{Judge} & \hcol{Acc} & \hcol{F1} & \hcol{Judge} & \hcol{Acc} & \hcol{F1} & \hcol{Judge} & \hcol{Acc} & \hcol{F1} & \hcol{Judge} \\
\Xhline{1pt}
\multirow{11}{*}{\mdl{Qwen3}{4B}} & \multicolumn{13}{c}{\cellcolor{catcolor}\textit{Non-collaborative}} \\
  & \quad \mname{Zero-shot} & 0.18 & 0.15 & 0.40 & 0.02 & 0.04 & 0.23 & 0.02 & 0.01 & 0.19 & 0.04 & 0.04 & 0.23 \\
  & \quad \mname{Single RAG} & 0.53 & 0.36 & 0.58 & 0.51 & 0.43 & 0.57 & 0.59 & 0.52 & \hlsecond{0.72} & 0.56 & 0.47 & 0.67 \\
 & \multicolumn{13}{c}{\cellcolor{catcolor}\textit{Text / structured sharing}} \\
  & \quad \mname{Dist.\ RAG} & \hlsecond{0.63} & \hlsecond{0.49} & \hlfirst{0.75} & \hlsecond{0.60} & \hlsecond{0.59} & 0.65 & \hlsecond{0.61} & \hlsecond{0.56} & \hlsecond{0.72} & \hlsecond{0.61} & \hlsecond{0.54} & \hlsecond{0.71} \\
  & \quad \mname{Global RAG} & \hlsecond{0.63} & 0.48 & \hlsecond{0.74} & 0.59 & \hlsecond{0.59} & 0.64 & \hlsecond{0.61} & 0.54 & \hlfirst{0.73} & \hlsecond{0.61} & 0.52 & \hlsecond{0.71} \\
  & \quad \mname{TextMAS} & 0.51 & 0.35 & 0.61 & 0.55 & 0.51 & 0.64 & 0.54 & 0.45 & 0.68 & 0.54 & 0.41 & 0.66 \\
 & \multicolumn{13}{c}{\cellcolor{catcolor}\textit{Latent communication}} \\
  & \quad \mname{Raw KV} & 0.60 & 0.48 & 0.70 & \hlfirst{0.62} & 0.55 & \hlsecond{0.67} & 0.57 & 0.52 & 0.69 & 0.58 & 0.50 & 0.69 \\
  & \quad \mname{LatentMAS} & 0.61 & 0.47 & 0.72 & 0.55 & 0.50 & 0.66 & 0.54 & 0.51 & 0.68 & 0.55 & 0.48 & 0.68 \\
  & \quad \method{} & \hlfirst{0.69} & \hlfirst{0.57} & \hlfirst{0.75} & \hlsecond{0.60} & \hlfirst{0.60} & \hlfirst{0.70} & \hlfirst{0.62} & \hlfirst{0.62} & \hlfirst{0.73} & \hlfirst{0.62} & \hlfirst{0.60} & \hlfirst{0.73} \\
\Xhline{0.6pt}
\multirow{11}{*}{\mdl{Llama-3.2}{3B}} & \multicolumn{13}{c}{\cellcolor{catcolor}\textit{Non-collaborative}} \\
  & \quad \mname{Zero-shot} & 0.05 & 0.02 & 0.35 & 0.02 & 0.02 & 0.13 & 0.01 & 0.00 & 0.13 & 0.02 & 0.01 & 0.16 \\
  & \quad \mname{Single RAG} & 0.40 & 0.32 & 0.51 & 0.42 & 0.35 & 0.49 & 0.46 & 0.38 & 0.60 & 0.44 & 0.35 & 0.57 \\
 & \multicolumn{13}{c}{\cellcolor{catcolor}\textit{Text / structured sharing}} \\
  & \quad \mname{Dist.\ RAG} & 0.47 & 0.33 & 0.58 & \hlsecond{0.53} & 0.47 & 0.60 & 0.50 & 0.42 & \hlsecond{0.64} & 0.50 & 0.41 & 0.62 \\
  & \quad \mname{Global RAG} & \hlsecond{0.51} & \hlsecond{0.41} & \hlsecond{0.61} & \hlfirst{0.55} & \hlsecond{0.48} & \hlfirst{0.63} & \hlsecond{0.53} & \hlsecond{0.44} & \hlsecond{0.64} & \hlsecond{0.53} & \hlsecond{0.42} & \hlsecond{0.63} \\
  & \quad \mname{TextMAS} & 0.30 & 0.19 & 0.54 & 0.26 & 0.29 & 0.56 & 0.35 & 0.26 & 0.59 & 0.32 & 0.25 & 0.57 \\
 & \multicolumn{13}{c}{\cellcolor{catcolor}\textit{Latent communication}} \\
  & \quad \mname{Raw KV} & 0.42 & 0.31 & 0.60 & 0.47 & 0.47 & 0.58 & 0.45 & 0.37 & \hlsecond{0.64} & 0.45 & 0.36 & 0.62 \\
  & \quad \mname{LatentMAS} & 0.32 & 0.23 & 0.53 & 0.38 & 0.40 & \hlsecond{0.62} & 0.41 & 0.35 & 0.63 & 0.39 & 0.33 & 0.61 \\
  & \quad \method{} & \hlfirst{0.72} & \hlfirst{0.54} & \hlfirst{0.72} & 0.51 & \hlfirst{0.59} & \hlsecond{0.62} & \hlfirst{0.67} & \hlfirst{0.66} & \hlfirst{0.76} & \hlfirst{0.64} & \hlfirst{0.62} & \hlfirst{0.72} \\
\Xhline{0.6pt}
\multirow{11}{*}{\mdl{MediPhi}{3.8B}} & \multicolumn{13}{c}{\cellcolor{catcolor}\textit{Non-collaborative}} \\
  & \quad \mname{Zero-shot} & 0.21 & 0.15 & 0.35 & 0.02 & 0.02 & 0.23 & 0.02 & 0.01 & 0.21 & 0.05 & 0.03 & 0.23 \\
  & \quad \mname{Single RAG} & 0.49 & 0.29 & 0.53 & 0.44 & 0.39 & 0.56 & 0.45 & 0.35 & 0.64 & 0.46 & 0.33 & 0.61 \\
 & \multicolumn{13}{c}{\cellcolor{catcolor}\textit{Text / structured sharing}} \\
  & \quad \mname{Dist.\ RAG} & 0.54 & \hlsecond{0.39} & 0.63 & 0.58 & \hlfirst{0.56} & \hlsecond{0.67} & \hlsecond{0.52} & \hlsecond{0.45} & 0.68 & \hlsecond{0.54} & \hlsecond{0.43} & 0.67 \\
  & \quad \mname{Global RAG} & 0.56 & \hlsecond{0.39} & \hlsecond{0.68} & \hlsecond{0.59} & \hlfirst{0.56} & \hlsecond{0.67} & \hlsecond{0.52} & 0.43 & \hlsecond{0.69} & \hlsecond{0.54} & 0.42 & \hlsecond{0.69} \\
  & \quad \mname{TextMAS} & 0.42 & 0.28 & 0.65 & 0.19 & 0.22 & 0.48 & 0.29 & 0.21 & 0.53 & 0.29 & 0.22 & 0.54 \\
 & \multicolumn{13}{c}{\cellcolor{catcolor}\textit{Latent communication}} \\
  & \quad \mname{Raw KV} & 0.56 & \hlsecond{0.39} & 0.63 & 0.52 & 0.53 & \hlsecond{0.67} & 0.49 & 0.42 & 0.66 & 0.51 & 0.41 & 0.66 \\
  & \quad \mname{LatentMAS} & \hlsecond{0.58} & \hlsecond{0.39} & 0.67 & 0.48 & 0.49 & \hlfirst{0.69} & 0.51 & 0.44 & \hlsecond{0.69} & 0.51 & 0.42 & \hlsecond{0.69} \\
  & \quad \method{} & \hlfirst{0.63} & \hlfirst{0.42} & \hlfirst{0.72} & \hlfirst{0.60} & \hlsecond{0.55} & \hlfirst{0.69} & \hlfirst{0.54} & \hlfirst{0.46} & \hlfirst{0.70} & \hlfirst{0.57} & \hlfirst{0.45} & \hlfirst{0.70} \\
\Xhline{1.2pt}
\end{tabular}}
\vspace{-0.3cm}
\end{table*}

\subsection{Skewed Hospital Partition}
\label{app:skew}

We further evaluate whether the results depend on balanced hospital-local databases. To create a non-uniform retrieval setting, we construct a disease-skewed Dirichlet partition of the training cases. Starting from the $7{,}220$ training cases used for hospital-local retrieval, we group cases by OMIM disease label. For each disease $d$ with $n_d$ cases, we sample a hospital allocation vector
\[
p_d \sim \mathrm{Dirichlet}(\alpha \mathbf{1}_N),
\]
where $N=5$ is the number of hospitals. We then shuffle the cases of disease $d$ with seed $42$ and assign them to hospitals according to $p_d$, with rounding remainders assigned to hospitals with the largest fractional parts. Smaller $\alpha$ produces stronger disease-level specialization. We use $\alpha=0.3$.

Table~\ref{tab:dirichlet-partition} summarizes the resulting partition. The split is substantially non-uniform at the disease level: the mean per-disease concentration, defined as the average maximum single-hospital share for each disease, is $0.65$, compared with $1/N=0.20$ for a uniform allocation. At the same time, total case counts remain moderately balanced, with case-count Gini $0.095$ and entropy $2.30$ bits (out of a maximum of $2.32$).

We evaluate the trained communication interface under this skewed retrieval setting. The test set and decoding protocol are unchanged. During evaluation, each hospital retrieves from the skewed hospital-local database instead of the balanced database, using the same cosine HPO retrieval procedure and $N=3$ hospital agents. This tests whether the learned latent communication interface remains useful when hospital evidence is distributed non-uniformly across retrieval pools.

Table~\ref{tab:skew-partition} reports per-cohort utility under the skewed partition. \method{} retains the strongest overall utility on both host backbones. On \llmname{Qwen3-4B}, it achieves the best overall accuracy and macro F1, with an overall judge score on par with the strongest baseline (\mname{Global RAG}). On \llmname{Llama-3.2-3B}, \method{} is best on all three metrics across every cohort by a large margin. These results indicate that the learned latent communication interface remains effective when hospital evidence is distributed non-uniformly across retrieval pools.

\begin{table}[t!]
\centering
\small
\caption{\textbf{Skewed Dirichlet hospital partition.} We use a disease-skewed Dirichlet allocation with $\alpha=0.3$ over $N=5$ hospitals. The partition is non-uniform at the disease level while keeping total case counts moderately balanced.}
\label{tab:dirichlet-partition}
\begin{tabular}{lrr}
\Xhline{1.2pt}
\rowcolor{headcolor}
\hgroup{Hospital} & \hgroup{\#Cases} & \hgroup{\#Unique OMIM} \\
\Xhline{1pt}
$H_0$ & 1{,}263 & 149 \\
$H_1$ & 1{,}149 & 143 \\
$H_2$ & 1{,}327 & 143 \\
$H_3$ & 1{,}799 & 166 \\
$H_4$ & 1{,}682 & 165 \\
\Xhline{1.2pt}
\end{tabular}
\end{table}

\begin{table*}[t!]
\centering
\small
\setlength{\tabcolsep}{4.5pt}
\renewcommand{\arraystretch}{1.08}
\caption{\textbf{Utility under a skewed Dirichlet hospital partition.} Per-cohort and overall accuracy (Acc), macro F1 (F1), and LLM-as-judge score (Judge) under the Dirichlet($\alpha{=}0.3$) partition. \hlfirst{Best} and \hlsecond{second best} method per column, ranked separately for each backbone; tied methods are both marked.}
\label{tab:skew-partition}
\vspace{-3pt}
\resizebox{\textwidth}{!}{%
\begin{tabular}{ll|ccc|ccc|ccc!{\vrule width 0.8pt}ccc}
\Xhline{1.2pt}
\rowcolor{headcolor}
  & & \multicolumn{3}{c|}{\hgroup{RareBench}}
  & \multicolumn{3}{c|}{\hgroup{Zenodo}}
  & \multicolumn{3}{c!{\vrule width 0.8pt}}{\hgroup{Phenopacket}}
  & \multicolumn{3}{c}{\hgroup{Overall}} \\
\rowcolor{headcolor}
\hgroup{Model} & \quad\hgroup{Method} & \hcol{Acc} & \hcol{F1} & \hcol{Judge} & \hcol{Acc} & \hcol{F1} & \hcol{Judge} & \hcol{Acc} & \hcol{F1} & \hcol{Judge} & \hcol{Acc} & \hcol{F1} & \hcol{Judge} \\
\Xhline{1pt}
\multirow{12}{*}{\llmname{Qwen3}} & \multicolumn{13}{c}{\cellcolor{catcolor}\textit{Non-collaborative}} \\
  & \quad \mname{Zero-shot} & 0.11 & 0.08 & 0.30 & 0.03 & 0.05 & 0.37 & 0.01 & 0.01 & 0.18 & 0.03 & 0.02 & 0.24 \\
  & \quad \mname{Single RAG} & 0.47 & 0.34 & 0.63 & 0.19 & 0.20 & 0.40 & 0.37 & 0.30 & 0.50 & 0.34 & 0.28 & 0.50 \\
 & \multicolumn{13}{c}{\cellcolor{catcolor}\textit{Text / structured sharing}} \\
  & \quad \mname{Dist.\ RAG} & 0.46 & 0.33 & 0.56 & 0.54 & 0.45 & 0.58 & 0.54 & 0.47 & 0.67 & 0.53 & 0.43 & 0.63 \\
  & \quad \mname{Global RAG} & \hlsecond{0.56} & \hlsecond{0.44} & \hlsecond{0.67} & \hlsecond{0.56} & \hlsecond{0.51} & 0.61 & 0.56 & \hlsecond{0.49} & \hlsecond{0.70} & \hlsecond{0.56} & \hlsecond{0.47} & \hlfirst{0.68} \\
  & \quad \mname{Struct.\ RAG} & 0.53 & 0.41 & 0.58 & 0.45 & 0.40 & 0.54 & \hlfirst{0.59} & \hlsecond{0.49} & \hlfirst{0.71} & 0.55 & 0.44 & 0.65 \\
  & \quad \mname{TextMAS} & 0.40 & 0.30 & 0.58 & 0.24 & 0.29 & 0.48 & 0.40 & 0.33 & 0.58 & 0.37 & 0.31 & 0.56 \\
 & \multicolumn{13}{c}{\cellcolor{catcolor}\textit{Latent communication}} \\
  & \quad \mname{Raw KV} & 0.47 & 0.38 & \hlsecond{0.67} & 0.52 & 0.41 & \hlsecond{0.62} & 0.48 & 0.40 & 0.65 & 0.49 & 0.39 & 0.64 \\
  & \quad \mname{LatentMAS} & 0.51 & 0.39 & 0.61 & 0.50 & 0.40 & 0.61 & 0.47 & 0.38 & 0.65 & 0.48 & 0.38 & 0.63 \\
  & \quad \method{} & \hlfirst{0.63} & \hlfirst{0.47} & \hlfirst{0.68} & \hlfirst{0.59} & \hlfirst{0.55} & \hlfirst{0.67} & \hlsecond{0.58} & \hlfirst{0.55} & 0.67 & \hlfirst{0.59} & \hlfirst{0.51} & \hlsecond{0.67} \\
\Xhline{1pt}
\multirow{12}{*}{\llmname{Llama-3.2}} & \multicolumn{13}{c}{\cellcolor{catcolor}\textit{Non-collaborative}} \\
  & \quad \mname{Zero-shot} & 0.02 & 0.01 & 0.21 & 0.02 & 0.02 & 0.09 & 0.00 & 0.00 & 0.10 & 0.01 & 0.00 & 0.11 \\
  & \quad \mname{Single RAG} & \hlsecond{0.40} & 0.25 & \hlsecond{0.56} & 0.08 & 0.13 & 0.27 & 0.29 & 0.21 & 0.44 & 0.26 & 0.19 & 0.42 \\
 & \multicolumn{13}{c}{\cellcolor{catcolor}\textit{Text / structured sharing}} \\
  & \quad \mname{Dist.\ RAG} & 0.35 & \hlsecond{0.33} & 0.47 & \hlsecond{0.33} & 0.30 & 0.50 & \hlsecond{0.47} & \hlsecond{0.40} & 0.61 & \hlsecond{0.42} & \hlsecond{0.36} & 0.57 \\
  & \quad \mname{Global RAG} & 0.23 & 0.19 & 0.39 & 0.28 & 0.33 & 0.51 & 0.44 & 0.36 & \hlsecond{0.62} & 0.38 & 0.32 & 0.56 \\
  & \quad \mname{Struct.\ RAG} & 0.35 & 0.23 & 0.46 & 0.31 & \hlsecond{0.34} & \hlsecond{0.66} & 0.39 & 0.31 & \hlsecond{0.62} & 0.37 & 0.28 & \hlsecond{0.61} \\
  & \quad \mname{TextMAS} & 0.30 & 0.20 & 0.46 & 0.21 & 0.23 & 0.34 & 0.26 & 0.19 & 0.41 & 0.25 & 0.19 & 0.40 \\
 & \multicolumn{13}{c}{\cellcolor{catcolor}\textit{Latent communication}} \\
  & \quad \mname{Raw KV} & 0.26 & 0.23 & 0.40 & 0.22 & 0.23 & 0.49 & 0.33 & 0.29 & 0.54 & 0.30 & 0.26 & 0.51 \\
  & \quad \mname{LatentMAS} & 0.26 & 0.23 & 0.42 & 0.21 & 0.28 & 0.50 & 0.32 & 0.29 & 0.55 & 0.29 & 0.27 & 0.52 \\
  & \quad \method{} & \hlfirst{0.68} & \hlfirst{0.50} & \hlfirst{0.72} & \hlfirst{0.66} & \hlfirst{0.59} & \hlfirst{0.71} & \hlfirst{0.72} & \hlfirst{0.71} & \hlfirst{0.77} & \hlfirst{0.70} & \hlfirst{0.65} & \hlfirst{0.75} \\
\Xhline{1.2pt}
\end{tabular}}
\end{table*}

\subsection{Bootstrap Confidence Intervals}
\label{app:bootstrap-ci}

All main results use greedy decoding with temperature $0$. Given fixed model weights, data splits, hospital partitions, retrieval indices, and prompts, inference is deterministic. We therefore estimate finite-test-set uncertainty using nonparametric bootstrap resampling over the $401$ test cases. For each bootstrap replicate, we resample test cases with replacement and recompute accuracy, macro F1, and LLM-as-judge score.

Table~\ref{tab:paired-bootstrap-ci} reports paired bootstrap confidence intervals for the difference between \method{} and the strongest non-\method{} baseline on each metric. On \llmname{Llama-3.2-3B}, \method{} significantly outperforms the strongest baseline on all three metrics. On \llmname{Qwen3-4B}, \method{} significantly improves macro F1, while accuracy and judge score are statistically comparable to the strongest baseline. This indicates that the \llmname{Qwen3-4B} gains are strongest for class-balanced performance, while the Llama gains are robust across all metrics.

\begin{table*}[t!]
\centering
\small
\setlength{\tabcolsep}{4.5pt}
\renewcommand{\arraystretch}{1.18}
\caption{\textbf{Paired bootstrap confidence intervals.} We report the difference between \method{} and the strongest non-\method{} baseline for each host and metric. Intervals are 95\% paired bootstrap confidence intervals over $1{,}000$ resamples of the $401$ test cases. A positive interval indicates that \method{} is significantly better under the bootstrap test.}
\label{tab:paired-bootstrap-ci}
\resizebox{\textwidth}{!}{%
\begin{tabular}{l|ccc}
\Xhline{1.2pt}
\rowcolor{headcolor}
\hgroup{Host} & \hcol{Acc} & \hcol{macro F1} & \hcol{Judge} \\
\Xhline{1pt}
\llmname{Qwen3}
& $+0.003$ $[-0.047,+0.050]$
& $+0.076$ $[+0.024,+0.125]$
& $+0.005$ $[-0.040,+0.050]$ \\
\llmname{Llama-3.2}
& $+0.262$ $[+0.212,+0.317]$
& $+0.254$ $[+0.202,+0.310]$
& $+0.100$ $[+0.047,+0.150]$ \\
\Xhline{1.2pt}
\end{tabular}}
\end{table*}

\section{Representation Analysis and Ablations}
\label{app:analysis}

\subsection{Trained Bottleneck Control}
\label{app:trained-bottleneck}

One concern is that \method{} trains a diagnosis-supervised latent interface, while \mname{Raw KV} and \mname{LatentMAS} are training-free latent baselines. To test whether supervised training alone explains the gains, we evaluate a length-matched trained bottleneck control, \mname{MLP-Pool32}. This baseline uses the same diagnosis cross-entropy objective and the same latent budget $m=32$, but replaces autoregressive latent distillation with a mean-pooled prompt representation followed by an MLP projection to $m$ latent positions. Table~\ref{tab:trained-bottleneck} shows that this trained control performs poorly on both host backbones. This suggests that the gains of \method{} do not come merely from adding trainable parameters or supervised diagnosis loss. Instead, preserving structured prompt-state information through autoregressive latent distillation is important for compact latent communication.

\begin{table}[t!]
\centering
\small
\setlength{\tabcolsep}{4.5pt}
\renewcommand{\arraystretch}{1.18}
\caption{\textbf{Trained bottleneck control.} \mname{MLP-Pool32} uses the same diagnosis supervision and latent budget as \method{}, but replaces autoregressive latent distillation with mean pooling followed by an MLP projection. \hlfirst{Best} per column pair.}
\label{tab:trained-bottleneck}
\resizebox{\columnwidth}{!}{%
\begin{tabular}{l|ccc|ccc}
\Xhline{1.2pt}
\rowcolor{headcolor}
 & \multicolumn{3}{c|}{\hgroup{\mname{MLP-Pool32}}} & \multicolumn{3}{c}{\hgroup{\method{}}} \\
\rowcolor{headcolor}
\hgroup{Host} & \hcol{Acc} & \hcol{F1} & \hcol{Judge} & \hcol{Acc} & \hcol{F1} & \hcol{Judge} \\
\Xhline{1pt}
\llmname{Qwen3}     & 0.03 & 0.02 & 0.31 & \hlfirst{0.64} & \hlfirst{0.61} & \hlfirst{0.72} \\
\llmname{Llama-3.2} & 0.05 & 0.05 & 0.30 & \hlfirst{0.73} & \hlfirst{0.68} & \hlfirst{0.77} \\
\Xhline{1.2pt}
\end{tabular}}
\end{table}

\subsection{Retrieval and Training-Data Sensitivity}
\label{app:ablations}

Figure~\ref{fig:analysis}(c) plots judge score against retrieval depth for three representative methods. Table~\ref{tab:depth-full} gives the full comparison across all metrics and every baseline. \method{} is strongest at each depth, and its scores change little as $k$ grows, while the text-sharing baselines improve with deeper retrieval. This is the expected pattern: the baselines consume retrieved cases as prompt context and benefit from more of it, whereas \method{} compresses whatever the hospital retrieves into a fixed-length block.

\begin{table}[t!]
\centering
\small
\setlength{\tabcolsep}{4.5pt}
\renewcommand{\arraystretch}{1.18}
\caption{\textbf{Retrieval depth.} Acc / macro F1 / Judge on the \llmname{Llama-3.2-3B} host as the number of cases retrieved per hospital varies. \hlfirst{Best} per column.}
\label{tab:depth-full}
\resizebox{\columnwidth}{!}{%
\begin{tabular}{l|ccc}
\Xhline{1.2pt}
\rowcolor{headcolor}
\hgroup{Method} & \hcol{Top-1} & \hcol{Top-2} & \hcol{Top-3} \\
\Xhline{1pt}
\mname{Dist.\ RAG}   & 0.45 / 0.40 / 0.63 & 0.54 / 0.50 / 0.66 & 0.55 / 0.50 / 0.64 \\
\mname{Global RAG}   & 0.47 / 0.42 / 0.62 & 0.51 / 0.47 / 0.66 & 0.52 / 0.46 / 0.69 \\
\mname{Struct.\ RAG} & 0.42 / 0.34 / 0.67 & 0.50 / 0.43 / 0.70 & 0.52 / 0.44 / 0.69 \\
\mname{TextMAS}      & 0.32 / 0.25 / 0.47 & 0.40 / 0.33 / 0.60 & 0.42 / 0.36 / 0.62 \\
\mname{Raw KV}       & 0.35 / 0.32 / 0.58 & 0.43 / 0.36 / 0.60 & 0.45 / 0.37 / 0.61 \\
\mname{LatentMAS}    & 0.33 / 0.31 / 0.59 & 0.33 / 0.30 / 0.57 & 0.37 / 0.31 / 0.59 \\
\method{}            & \hlfirst{0.73 / 0.68 / 0.77} & \hlfirst{0.74 / 0.68 / 0.78} & \hlfirst{0.74 / 0.67 / 0.79} \\
\Xhline{1.2pt}
\end{tabular}}
\end{table}

Figure~\ref{fig:analysis}(d) varies the fraction of the training split used to fit the latent interface. Validation and test sets are unchanged. At $25\%$, $50\%$, and $100\%$ of the training split, respectively, accuracy / macro F1 / judge score is $0.53/0.48/0.64$, $0.66/0.59/0.71$, and $0.73/0.68/0.77$. Performance rises with more data and has not saturated at the full split; at $25\%$ the interface already exceeds every comparison method in accuracy and macro F1 under the same top-1 retrieval setting.

\subsection{Component Design}
\label{app:component-ablation}

We further isolate three interface choices on \llmname{Llama-3.2-3B}. The distiller and boundary-embedding studies use the default same-backbone configuration; the aggregation study uses the heterogeneous Degree~1 L:Q:L configuration, with latent-block size, training data, and optimization held fixed. Results are comparable only within each component block in Table~\ref{tab:component-ablation}.

\begin{table}[t!]
\centering
\small
\setlength{\tabcolsep}{4.5pt}
\renewcommand{\arraystretch}{1.18}
\caption{\textbf{Component ablations.} Distiller-architecture and boundary-embedding results use the default same-backbone \method{} setting with a \llmname{Llama-3.2-3B} host. Host-side aggregation is evaluated under the heterogeneous Deg1 configuration (L:Q:L), with latent-block size, training data, and optimization held fixed. Scores should be compared only within each block. Best values are highlighted, with ties marked jointly.}
\label{tab:component-ablation}
\resizebox{\columnwidth}{!}{%
\begin{tabular}{l|ccc}
\Xhline{1.2pt}
\rowcolor{headcolor}
\hgroup{Variant} & \hcol{Acc} & \hcol{Macro F1} & \hcol{Judge} \\
\Xhline{1pt}
\multicolumn{4}{c}{\cellcolor{catcolor}\textit{Distiller architecture}} \\
\mname{Linear} & 0.72 & 0.68 & 0.75 \\
\mname{LN + Linear} (ours) & 0.73 & 0.68 & 0.77 \\
\mname{2-layer MLP} (\mname{Linear}$\rightarrow$\mname{GELU}$\rightarrow$\mname{Linear})
  & \hlfirst{0.74} & \hlfirst{0.69} & \hlfirst{0.79} \\
\Xhline{0.6pt}
\multicolumn{4}{c}{\cellcolor{catcolor}\textit{Boundary embeddings}} \\
\mname{Off} & \hlfirst{0.73} & \hlfirst{0.68} & 0.76 \\
\mname{On} (ours) & \hlfirst{0.73} & \hlfirst{0.68} & \hlfirst{0.77} \\
\Xhline{0.6pt}
\multicolumn{4}{c}{\cellcolor{catcolor}\textit{Host-side aggregation}} \\
\mname{Mean pooling} & 0.52 & 0.44 & 0.60 \\
\mname{Concatenation} (ours) & \hlfirst{0.72} & \hlfirst{0.66} & \hlfirst{0.76} \\
\Xhline{1.2pt}
\end{tabular}}
\end{table}

The two-layer MLP yields only marginal gains over LayerNorm followed by one linear projection, so we retain the lighter default distiller. Removing the boundary embeddings leaves accuracy and macro F1 unchanged and changes judge score by only $0.01$, consistent with their intended role of delimiting hospital-specific blocks. Concatenation substantially outperforms position-wise mean pooling; this result is consistent with preserving each hospital's block as a distinct segment for host attention instead of collapsing the hospital-specific evidence.

\subsection{Latent Disease Structure}
\label{app:tsne}

\begin{figure}[t!]
\centering
\includegraphics[width=\columnwidth]{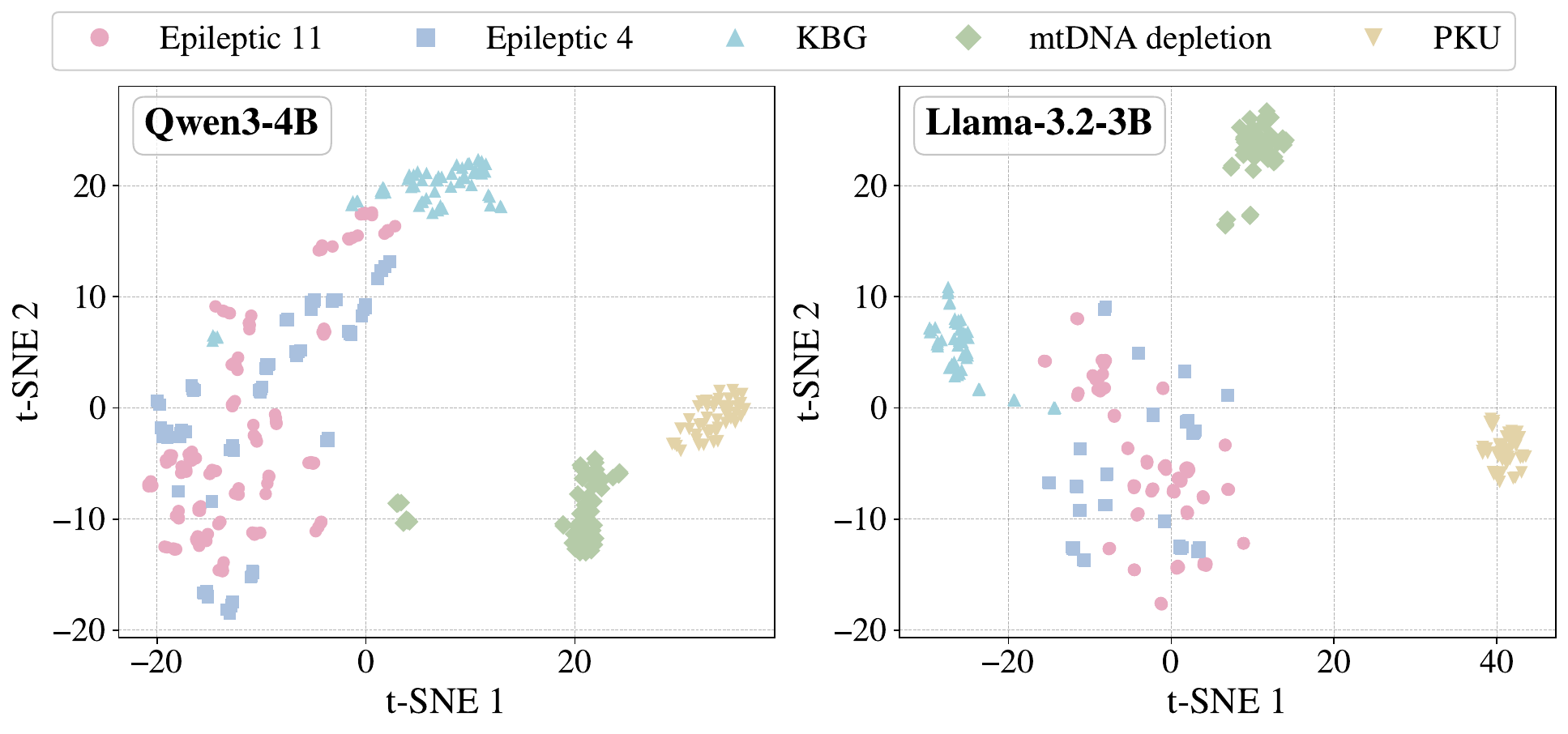}
\caption{\textbf{Disease structure in latent space.} t-SNE of compact latent blocks from two encoder families.}
\label{fig:tsne}
\end{figure}

Figure~\ref{fig:tsne} visualises compact latent blocks from two encoder families with t-SNE. Several diseases form separated clusters within each family, suggesting that the blocks retain disease-discriminative structure after compression. Related subtypes partially overlap, which is consistent with diagnostic similarity rather than arbitrary label separation. The two panels are computed independently per family and do not imply that the two latent spaces are aligned.

\section{Privacy and Leakage}
\label{app:privacy}

\subsection{Additional Reconstruction Leakage Results}
\label{app:leakage}

\paragraph{Attack and metric.}
We reuse the frozen-LLM continuation attack from \S\ref{sec:problem-setup}: a passive observer intercepts the transmitted object $\mathcal{T}_{H_i}$, injects it as past key--value states into the frozen LLM that produced it, prompts the model to repeat the previous hospital case content, and greedily decodes a reconstruction $\tilde{x}_i$. We score $\tilde{x}_i$ against the source local prompt $x_i$ with BERTScore. Because raw \llmname{RoBERTa-large} BERTScore remains high even for unrelated clinical sentences, around $0.85$ for random sentence pairs, it can reflect shared medical phrasing rather than successful reconstruction. We therefore report \emph{rescaled} BERTScore, which subtracts a random-sentence baseline so that $1$ corresponds to full reconstruction, $0$ to random-baseline similarity, and negative values to reconstructions less similar to the source than random text. Each cell averages $n=2{,}005$ prompt pairs, corresponding to $401$ test cases across five hospital databases.

\begin{table}[t!]
\centering
\small
\setlength{\tabcolsep}{4.5pt}
\renewcommand{\arraystretch}{1.18}
\caption{\textbf{Rescaled BERTScore for KV reconstruction leakage.} Lower is better ($n=2{,}005$ prompt pairs per cell). BERTScore is rescaled against a random-sentence baseline so that $1$ means full reconstruction, $0$ means indistinguishable from random text, and negative values mean less similar than random. \hlfirst{Best} per row.}
\label{tab:leakage-bertscore}
\vspace{-3pt}
\resizebox{\columnwidth}{!}{%
\begin{tabular}{l|ccc}
\Xhline{1.2pt}
\rowcolor{headcolor}
\hgroup{Model} & \hcol{\mname{Raw KV}} & \hcol{\mname{LatentMAS}} & \hcol{\method{}} \\
\Xhline{1pt}
\llmname{Qwen3}     & 0.593 & 0.593 & \hlfirst{$-$0.125} \\
\llmname{Llama-3.2} & 0.673 & 0.812 & \hlfirst{$-$0.322} \\
\llmname{MediPhi}   & 0.724 & 0.712 & \hlfirst{$-$0.308} \\
\Xhline{1.2pt}
\end{tabular}}
\end{table}

\paragraph{Results.}
Table~\ref{tab:leakage-bertscore} shows that both raw-latent baselines leak substantial source content. \mname{Raw KV} reaches rescaled BERTScore between $0.59$ and $0.72$, since it transmits the full prompt-length KV cache $\mathcal{M}_{H_i}$ and enables the model to largely paraphrase the original prompt. \mname{LatentMAS} leaks at a comparable level, between $0.59$ and $0.81$, showing that latent communication alone does not remove surface-level leakage. In contrast, \method{} attains negative rescaled BERTScore on all three backbones, ranging from $-0.125$ to $-0.322$. The reconstruction from its compact latent block $\mathcal{Z}_{H_i}$ is less similar to the source than random text, indicating little recoverable case-specific content from the intercepted block. This suggests that the leakage reduction of \method{} comes from compressing the prompt-length KV cache into a compact distilled latent block, not from latent communication alone.

\subsection{Attribute Leakage under Prompt-Based Extraction Attacks}
\label{app:prompt-extraction}

Token F1 and BERTScore measure whether the attacker can recover text similar to the source local prompt. We further evaluate whether the reconstructed outputs reveal clinically salient attributes from the retrieved hospital case. Specifically, we measure \emph{disease leakage}, the fraction of cases where the retrieved disease label is recovered, and \emph{phenotype recall}, the fraction of retrieved phenotype terms recovered from the source local prompt. Lower values indicate less attribute-level clinical-content leakage.

We evaluate two prompt-based extraction attacks. The first is the verbatim continuation attack used in Sec.~\ref{sec:problem-setup}. The second is a stronger template-aware attack: the attacker is given the hospital-case template and is explicitly asked to recover the retrieved disease label and phenotype terms. In both attacks, the attacker injects the intercepted transmitted state into the frozen LLM that produced it. No attacker model is trained.

\begin{table*}[t!]
\centering
\small
\setlength{\tabcolsep}{8pt}
\renewcommand{\arraystretch}{1.18}
\caption{\textbf{Attribute-level leakage under prompt-based extraction attacks.} Disease recall is the fraction of cases in which the retrieved disease label appears in the attacker's continuation. Phenotype measures recall of retrieved phenotype terms. The template-aware attack reveals the hospital-case template and explicitly requests the disease and phenotype fields. Lower values indicate less attribute-level leakage; \hlfirst{Best} method per column, ranked separately for each backbone.}
\label{tab:attribute-leakage}
\resizebox{0.78\textwidth}{!}{%
\begin{tabular}{ll|cc|cc}
\Xhline{1.2pt}
\rowcolor{headcolor}
 & & \multicolumn{2}{c|}{\hgroup{Verbatim}} & \multicolumn{2}{c}{\hgroup{Template-aware}} \\
\rowcolor{headcolor}
\hgroup{Model} & \hgroup{Method} & \hcol{Disease recall $\downarrow$} & \hcol{Phenotype $\downarrow$} & \hcol{Disease recall $\downarrow$} & \hcol{Phenotype $\downarrow$} \\
\Xhline{1pt}
\multirow{3}{*}{\llmname{Qwen3}}
& \mname{Raw KV}    & 1.00 & 1.00 & 0.89 & 0.86 \\
& \mname{LatentMAS} & 1.00 & 1.00 & 0.90 & 0.85 \\
& \method{}         & \hlfirst{0.02} & \hlfirst{0.01} & \hlfirst{0.20} & \hlfirst{0.14} \\
\Xhline{0.6pt}
\multirow{3}{*}{\llmname{Llama-3.2}}
& \mname{Raw KV}    & 0.99 & 1.00 & 0.88 & 0.97 \\
& \mname{LatentMAS} & 0.98 & 0.99 & 0.75 & 0.89 \\
& \method{}         & \hlfirst{0.14} & \hlfirst{0.03} & \hlfirst{0.16} & \hlfirst{0.03} \\
\Xhline{0.6pt}
\multirow{3}{*}{\llmname{MediPhi}}
& \mname{Raw KV}    & 0.99 & 1.00 & 0.76 & 0.94 \\
& \mname{LatentMAS} & 1.00 & 1.00 & 0.77 & 0.96 \\
& \method{}         & \hlfirst{0.04} & \hlfirst{0.03} & \hlfirst{0.05} & \hlfirst{0.04} \\
\Xhline{1.2pt}
\end{tabular}}
\end{table*}

Table~\ref{tab:attribute-leakage} shows that \mname{Raw KV} and \mname{LatentMAS} reveal most retrieved disease labels and phenotype terms under both prompt variants. The template-aware attack is stronger than the verbatim prompt, and increases leakage for \method{} on \llmname{Qwen3-4B}. However, \method{} still remains far below the raw-latent baselines: even in the highest-leakage case, disease leakage is $0.20$ versus $0.89$--$0.90$, and phenotype recall is $0.14$ versus $0.85$--$0.86$. Across backbones, \method{} keeps disease leakage at or below $0.20$ and phenotype recall at or below $0.14$. These results indicate that the leakage reduction is not only a text-overlap effect, and that even a template-aware attacker recovers substantially less clinical content from compact latent blocks.

\subsection{Learned Attacks and Full Evaluation Matrix}
\label{app:leakage-attacks}

\paragraph{Side information and controls.}
We evaluate three learned attacks on same-backbone \methodh{} representations produced by each of \llmname{Qwen3-4B}, \llmname{Llama-3.2-3B}, and \llmname{MediPhi}, together with the corresponding baselines. The attacker passively observes the transmitted representation, knows the participating backbone, prompt template, and retrieval procedure, and may train on auxiliary examples from the same distribution, but it has no access to the target hospital database or the target retrieved record in clear text at test time. In addition to the transmission, the probe and inversion decoder receive the broadcast query HPO set, while membership inference receives a candidate case and target-hospital identifier. The no-transmission control for each attack receives the same side information but no transmitted representation. It therefore provides a reference floor for the incremental recoverability attributable to transmission; the table entries remain absolute metric values, and Reduction is computed directly from \mname{Raw KV} to \method{}.

\paragraph{Shared representation encoder.}
All learned attacks use the same representation-encoder design across communication methods, while each backbone--method--attack combination is trained independently from scratch. At each backbone layer, key and value states are concatenated and mapped through a learned layer-specific bottleneck. The resulting position representations are combined across layers, projected to a common width, augmented with learned positional embeddings and a \texttt{[CLS]} token, and processed by a pre-LN Transformer encoder. Side information is encoded as an additional token after \texttt{[CLS]}. The attribute probe and membership classifier use the \texttt{[CLS]} state, whereas the inversion decoder cross-attends to the full encoded sequence. The encoder supports the complete intercepted object for each method: up to $320$ positions for \mname{Raw KV}, up to $322$ for \mname{LatentMAS}, and exactly $34$ for \methodh{}.

For \methodh{}, the intercepted sequence contains every position sent by the hospital. After encoding its private prompt, the hospital appends $\langle\textsc{bop}\rangle$, autoregressively generates $m=32$ latent positions, appends $\langle\textsc{eop}\rangle$, and transmits the KV states of these final $m+2=34$ positions. The two boundary positions are not public constants: their KV states are computed in prompt context, with the begin boundary conditioned on the private prompt and the end boundary additionally conditioned on the generated latent sequence. We expose all $34$ positions to the attacker; withholding the boundary states would understate the transmitted attack surface.

\paragraph{Attribute probe.}
Linear readout heads on the pooled representation predict the retrieved case's OMIM class and HPO terms. We train the disease head with softmax cross-entropy and the phenotype head with frequency-weighted binary cross-entropy. The probe inherits the case-level train, validation, and test partitions, so evaluation queries are disjoint from attacker training. Checkpoints and multilabel decision thresholds are selected only on validation data and then fixed for test evaluation. We report OMIM exact match and HPO micro-F1.

\paragraph{Autoregressive inversion.}
A Transformer encoder--decoder reconstructs a serialized record by cross-attending to the full encoded transmission. Its target contains only the retrieved OMIM identifier and HPO terms; query fields and template boilerplate are excluded so that overlap cannot be obtained from shared scaffolding. We train with teacher forcing and an up-weighted end-of-sequence term, restore the best validation checkpoint, and use greedy decoding at test time. We evaluate the decoded identifier by OMIM exact match and the recovered phenotype terms by HPO micro-F1.

\paragraph{Retrieval-database membership inference.}
Here the secret is whether a candidate case belongs to a hospital's retrieval database. For each candidate, we construct one positive example with its member hospital and one hard negative with the non-member hospital that assigns the highest retrieval score. Candidate content is fixed within the pair, while the target hospital and its intercepted transmission vary. We split candidates $60/20/20$ into train, validation, and test sets, stratified by member hospital and disjoint by case identifier; both examples for a candidate remain in the same split. A binary classifier operates on the pooled transmission representation together with candidate and hospital embeddings. We select checkpoints by validation ROC-AUC and report test TPR at $1\%$ FPR.

\paragraph{Training and reporting.}
We optimize every attacker with AdamW, linear warmup followed by decay, gradient clipping, early stopping, and restoration of the best validation checkpoint. No attacker weights are shared across communication methods or backbones. Each backbone--method--attack combination, including its no-transmission control, is trained with three random seeds, and Tables~\ref{tab:privacy-main} and~\ref{tab:leakage-full} report the mean across seeds.

Table~\ref{tab:privacy-main} reports the full attack ladder on \llmname{Qwen3-4B} and \llmname{Llama-3.2-3B}. Table~\ref{tab:leakage-full} extends this matrix with \llmname{MediPhi}, the unrescaled BERTScore row, and HPO micro-F1 for the learned probe and inversion decoder. Across the three backbones, \method{} reduces OMIM exact-match leakage by $58$--$67\%$, HPO micro-F1 by $58$--$76\%$, and membership-inference TPR at $1\%$ FPR by $83$--$97\%$ relative to \mname{Raw KV}. Its membership-inference TPR remains between $0.02$ and $0.17$, compared with $0.79$--$1.00$ for \mname{Raw KV}. Together with the training-free attacks, these results show a consistent empirical reduction in recoverable clinical information across all three evaluated backbones.

\begin{table*}[t!]
\centering
\small
\setlength{\tabcolsep}{4pt}
\renewcommand{\arraystretch}{1.18}
\caption{\textbf{Leakage across all evaluated backbone--attack pairs.} This table extends Table~\ref{tab:privacy-main} with \llmname{MediPhi}, the unrescaled BERTScore row, and the HPO micro-F1 scores of the two learned attacks. Lower is better. No~tx. denotes an attack-specific no-transmission control: the attacker receives no transmitted representation. It serves as the control baseline for measuring additional leakage attributable to transmission. Learned-attack entries are means over three random seeds. \colorbox{firstcolor}{\strut Reduction} is the relative drop from \mname{Raw KV} to \method{}. Metrics are not comparable across rows.}
\label{tab:leakage-full}
\vspace{-3pt}
\resizebox{0.72\textwidth}{!}{%
\begin{tabular}{ll|cccc|c}
\Xhline{1.2pt}
\rowcolor{headcolor}
\hgroup{Attack} & \hgroup{Metric} & \hcol{No tx.} & \hcol{\mname{Raw KV}} & \hcol{\mname{LatentMAS}} & \hcol{\method{}} & \hgroup{Reduction} \\
\Xhline{1pt}
\multicolumn{7}{c}{\cellcolor{catcolor}\hgroup{\llmname{Qwen3-4B}}} \\
Prompt reconstruction     & Token F1 $\downarrow$       & 0.05 & 0.60 & 0.60 & 0.07 & \reduc{$\downarrow$88\%} \\
Prompt reconstruction     & BERTScore $\downarrow$      & 0.78 & 0.93 & 0.93 & 0.81 & \reduc{$\downarrow$13\%} \\
Template-aware extraction & Disease recall $\downarrow$ & 0.00 & 0.89 & 0.90 & 0.20 & \reduc{$\downarrow$78\%} \\
Attribute probe           & OMIM exact $\downarrow$      & 0.17 & 0.62 & 0.62 & 0.23 & \reduc{$\downarrow$63\%} \\
Attribute probe           & HPO micro-F1 $\downarrow$    & 0.05 & 0.33 & 0.33 & 0.14 & \reduc{$\downarrow$58\%} \\
Inversion decoder         & OMIM exact $\downarrow$      & 0.27 & 0.97 & 0.96 & 0.40 & \reduc{$\downarrow$59\%} \\
Inversion decoder         & HPO micro-F1 $\downarrow$    & 0.21 & 0.90 & 0.90 & 0.22 & \reduc{$\downarrow$76\%} \\
Membership inference      & TPR@1\%FPR $\downarrow$      & 0.00 & 0.99 & 0.93 & 0.17 & \reduc{$\downarrow$83\%} \\
\Xhline{0.6pt}
\multicolumn{7}{c}{\cellcolor{catcolor}\hgroup{\llmname{Llama-3.2-3B}}} \\
Prompt reconstruction     & Token F1 $\downarrow$       & 0.07 & 0.68 & 0.82 & 0.07 & \reduc{$\downarrow$90\%} \\
Prompt reconstruction     & BERTScore $\downarrow$      & 0.78 & 0.95 & 0.97 & 0.78 & \reduc{$\downarrow$18\%} \\
Template-aware extraction & Disease recall $\downarrow$ & 0.00 & 0.88 & 0.75 & 0.16 & \reduc{$\downarrow$82\%} \\
Attribute probe           & OMIM exact $\downarrow$      & 0.17 & 0.61 & 0.60 & 0.20 & \reduc{$\downarrow$67\%} \\
Attribute probe           & HPO micro-F1 $\downarrow$    & 0.05 & 0.32 & 0.32 & 0.13 & \reduc{$\downarrow$59\%} \\
Inversion decoder         & OMIM exact $\downarrow$      & 0.27 & 0.99 & 0.99 & 0.38 & \reduc{$\downarrow$62\%} \\
Inversion decoder         & HPO micro-F1 $\downarrow$    & 0.23 & 0.87 & 0.87 & 0.27 & \reduc{$\downarrow$69\%} \\
Membership inference      & TPR@1\%FPR $\downarrow$      & 0.00 & 1.00 & 0.83 & 0.03 & \reduc{$\downarrow$97\%} \\
\Xhline{0.6pt}
\multicolumn{7}{c}{\cellcolor{catcolor}\hgroup{\llmname{MediPhi}}} \\
Prompt reconstruction     & Token F1 $\downarrow$       & 0.07 & 0.76 & 0.75 & 0.07 & \reduc{$\downarrow$91\%} \\
Prompt reconstruction     & BERTScore $\downarrow$      & 0.77 & 0.95 & 0.95 & 0.78 & \reduc{$\downarrow$18\%} \\
Template-aware extraction & Disease recall $\downarrow$ & 0.00 & 0.76 & 0.77 & 0.05 & \reduc{$\downarrow$93\%} \\
Attribute probe           & OMIM exact $\downarrow$      & 0.17 & 0.55 & 0.62 & 0.22 & \reduc{$\downarrow$60\%} \\
Attribute probe           & HPO micro-F1 $\downarrow$    & 0.05 & 0.30 & 0.31 & 0.10 & \reduc{$\downarrow$67\%} \\
Inversion decoder         & OMIM exact $\downarrow$      & 0.27 & 0.95 & 0.93 & 0.40 & \reduc{$\downarrow$58\%} \\
Inversion decoder         & HPO micro-F1 $\downarrow$    & 0.24 & 0.84 & 0.84 & 0.25 & \reduc{$\downarrow$70\%} \\
Membership inference      & TPR@1\%FPR $\downarrow$      & 0.00 & 0.79 & 0.94 & 0.02 & \reduc{$\downarrow$97\%} \\
\Xhline{1.2pt}
\end{tabular}}
\end{table*}

\section{LLM Usage}
\label{app:llm-usage}

We used Large Language Models (ChatGPT/Claude/Gemini) for grammatical correction in this manuscript. Separately, \llmname{GPT-5.5} generated the paired clinical narratives in \benchnametext{} from HPO term names under the constrained protocol described in Appendix~\ref{app:note-corpus}; it received no disease name, OMIM identifier, or causal gene, and the outputs were screened for explicit diagnostic-label leakage. LLMs played no role in research ideation or methodology. All technical contributions, analyses, and scientific conclusions are the authors' own.

\end{document}